\documentclass{article}
\usepackage{enumitem} %
\usepackage{subcaption}
\usepackage{booktabs}
\usepackage{svg}
\usepackage{tikz}
\usepackage{graphicx}  
\usepackage{footnote}
\usepackage{makecell} 
\usepackage{adjustbox}
\usepackage{amsmath}
\usepackage{caption}
\usepackage[accepted]{icml2025}
\usepackage{float}
\usepackage{array}
\usepackage{afterpage}
\usetikzlibrary{calc}
\usepackage{tabularx}
\usepackage{amssymb}
\usepackage{xcolor}
\usepackage{tcolorbox}
\usepackage{lipsum} 
\usepackage{multirow}
\usepackage{hyperref}

\title{Multi-Head Explainer: A General Scalable Framework to Improve Explainability in CNNs and Transformers}

\author{
  Bohang Sun \\
  School of Information and Software Engineering \\
  University of Electronic Science and Technology of China \\
  \texttt{bobsun@std.uestc.edu.cn} \\
  \and
  Pietro Liò \\
  Computer Science Department \\
  University of Cambridge \\
  \texttt{pl219@cam.ac.uk}
}

\begin{document}

\maketitle



\begin{abstract}
In this study, we introduce the Multi-Head Explainer (MHEX), a versatile and modular framework that enhances both the explainability and accuracy of Convolutional Neural Networks (CNNs) and Transformer-based models. MHEX consists of three core components: an Attention Gate that dynamically highlights task-relevant features, Deep Supervision that guides early layers to capture fine-grained details pertinent to the target class, and an Equivalent Matrix that unifies refined local and global representations to generate comprehensive saliency maps. Our approach demonstrates superior compatibility, enabling effortless integration into existing residual networks like ResNet and Transformer architectures such as BERT with minimal modifications. Extensive experiments on benchmark datasets in medical imaging and text classification show that MHEX not only improves classification accuracy but also produces highly interpretable and detailed saliency scores.\footnote{For those who wish to explore the implementation of MHEX, we kindly refer you to the guidelines in Appendix~\ref{appendix:Guidelines}}

\end{abstract}

\section{Introduction}

Deep learning models have achieved remarkable success across various fields, but their complex and opaque nature often hinders interpretability and trust, particularly in critical applications.

In the realm of computer vision, medical imaging stands out as a field where the need for explainable models is particularly pronounced. Medical professionals rely on precise and interpretable model predictions to make informed decisions, as inaccuracies or lack of transparency can have serious consequences for patient care \cite{shamshad2023transformers}. However, standard explainability methods such as Grad-CAM \cite{selvaraju2017grad} and SHAP \cite{lundberg2017unified} often fall short in this context. These methods frequently fail to capture the rich texture details and subtle features inherent in medical images, leading to incomplete or misleading interpretations that are insufficient for clinical use.

Similarly, in Transformer-based architectures, which have become the backbone of many state-of-the-art NLP models \cite{vaswani2017attention}, significant challenges persist regarding explainability. Recent studies have highlighted issues such as over-smoothing \cite{dovonon2024settingrecordstraighttransformer}, where attention mechanisms tend to produce uniform attention distributions across layers, diluting the interpretative power of attention scores. Additionally, research by Jain and Wallace \cite{jain2019attentionexplanation} has demonstrated that attention weights in Transformers do not necessarily provide meaningful explanations for model predictions.

To address these challenges, we introduce \textbf{Multi-Head Explainer (MHEX)}\footnote{The code is available at \url{https://github.com/BobSun98/Deep-Explain}.}, a general and modular framework designed to enhance the explainability and accuracy of both Convolutional Neural Networks (CNNs) and Transformer-based models. MHEX serves as a versatile "scaffold" that can be seamlessly integrated into various network architectures, including residual networks like ResNet \cite{he2015deepresiduallearningimage} and Transformer models such as BERT \cite{DBLP:journals/corr/abs-1810-04805}. After the model is fine-tuned, the MHEX prediction heads modules can be removed, ensuring minimal impact on the original model's architecture and performance.
Our primary contributions are threefold:
\begin{enumerate}[left=0pt]
    \item \textbf{MHEX Framework}: We present the Multi-Head Explainer framework, detailing its components and demonstrating its application in enhancing both residual networks and Transformer architectures. This framework significantly improves the generation of comprehensive and interpretable saliency scores across different model types.
    \item \textbf{Method Leveraging Non-Negativity}: We introduce a novel method that leverages non-negativity constraints induced by activation functions such as ReLU to reduce noise in saliency maps. This approach enhances the clarity and focus of feature importance.
    \item \textbf{Tailored Metrics}: We design specialized evaluation metrics tailored to assess the quality of saliency maps in complex domains. These metrics provide a more nuanced and accurate evaluation of model explanations.
\end{enumerate}


\section{Related Work}
\subsection{Explainability in CNNs}

Convolutional Neural Networks (CNNs) have been extensively utilized across various domains. Over time, CNN architectures have evolved to incorporate sophisticated mechanisms that enhance their representational power and adaptability. For instance, attention-based CNNs such as Squeeze-and-Excitation Networks (SENet) \cite{hu2018squeeze}, Convolutional Block Attention Module (CBAM) \cite{woo2018cbam}, and Bottleneck Attention Module (BAM) \cite{park2018bam} dynamically emphasize important features, thereby improving both performance and interpretability. Additionally, spherical CNNs \cite{cohen2018sphericalcnns} have been developed to process 360-degree spherical data, maintaining rotation and translation invariance on spherical surfaces, which is crucial for applications like omnidirectional vision and geospatial analysis. Recent hybrid architectures, such as ConvNeXt \cite{Liu_2022_CVPR} and MobileViT \cite{mehta2021mobilevit}, integrate Transformer within CNN frameworks to leverage both local feature extraction and global contextual understanding.

Despite the advancements in CNN architectures, the interpretability of these models remains a critical concern. Numerous saliency-based methods have been proposed to address this issue, aiming to highlight important regions in input images that contribute to model predictions. Techniques like Grad-CAM \cite{selvaraju2017grad}, Grad-CAM++ \cite{chattopadhay2018grad}, Layer-CAM \cite{jiang2021layercam}, Score-CAM \cite{wang2020score}, Eigen-CAM \cite{9206626}, SHAP \cite{lundberg2017unified}, Integrated Gradients \cite{sundararajan2017axiomatic}, and Guided Backpropagation \cite{springenberg2014striving} have been widely adopted to visualize model decision-making processes. However, these methods often struggle to capture the rich texture details and subtle features inherent in medical images, leading to incomplete or misleading interpretations.

\subsection{Transformer Explainability}

Transformer-based architectures have revolutionized natural language processing (NLP) and have made significant inroads into computer vision tasks \cite{vaswani2017attention, DBLP:journals/corr/abs-1810-04805}. Despite their success, the interpretability of Transformers remains a challenging issue. Early efforts focused on analyzing attention weights to understand model focus \cite{clark2019doesbertlookat}, but subsequent studies revealed that these weights often exhibit intricate and repetitive patterns that do not necessarily correlate with meaningful explanations \cite{kovaleva2019revealingdarksecretsbert}.

Recent advancements have sought to develop more sophisticated explainability techniques tailored for Transformers. For instance, \citet{qiang2022attcat} introduced AttCat, which leverages attentive class activation tokens by integrating encoded features, gradients, and self-attention weights to provide more granular and faithful explanations of Transformer predictions. Similarly, ViT-CX \cite{xie2023vitcxcausalexplanationvision} focuses on causal explanations in Vision Transformers, enhancing the interpretability of models in computer vision by identifying causal relationships between input features and model outputs. Attention Flow \cite{abnar2020quantifying} proposed methods such as attention rollout and attention flow to model information flow in Transformers using a directed acyclic graph (DAG), offering more accurate and reliable quantifications of self-attention mechanisms compared to raw attention weights.

\section{Method}

In this section, we present the \textbf{Multi-Head Explainer (MHEX)} framework, its core components, and its integration into both CNNs and Transformer-based models, along with neuron analysis techniques and evaluation metrics for saliency maps.

\subsection{Overall Framework Introduction}

Multi-Head Explainer (MHEX) is a modular framework designed to enhance the explainability and accuracy of deep learning models, including CNNs and Transformers. Its design allows seamless integration into various architectures, improving model interpretability across tasks. The MHEX framework comprises three components:
\begin{itemize}[left=0pt, itemsep=0pt, topsep=0pt]
    \item \textbf{Attention Gate}: Dynamically emphasizes task-relevant features.
    \item \textbf{Deep Supervision}: Guides early layers to capture fine-grained features pertinent to the target class.
    \item \textbf{Equivalent Matrix}: Unifies refined local and global representations to generate comprehensive saliency scores.
\end{itemize}

By integrating these components, MHEX enhances the model's ability to produce detailed and interpretable saliency scores, thereby bridging the gap between model performance and interpretability.

\begin{figure}[h!]
    \centering
    \includegraphics[width=1\linewidth]{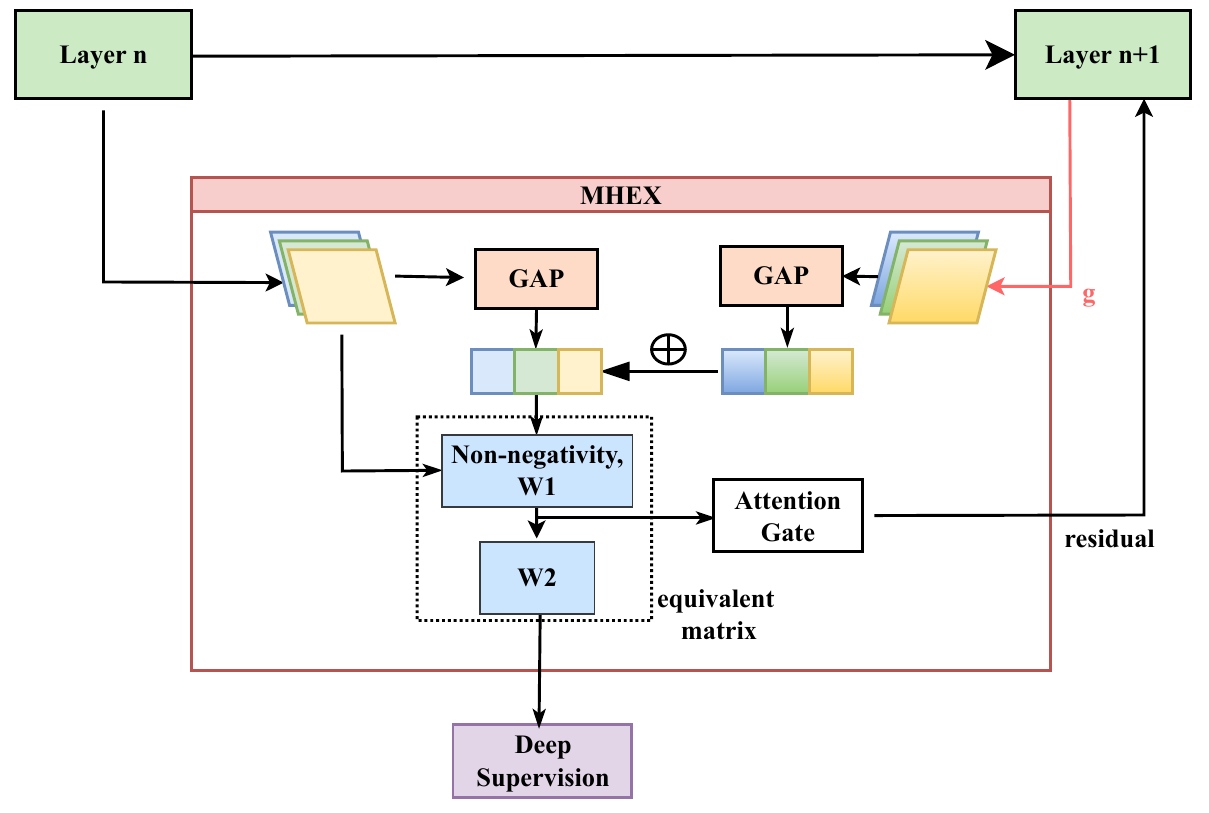}
    \caption{The internal structure of the Multi-Head Explainer (MHEX)}
    \label{fig:den_resnet}
\end{figure}

\subsection{MHEX Core Components}
\subsubsection{Attention Gate}
\label{attentiongate}
The \textbf{Attention Gate}~\cite{schlemper2019attention} prioritizes task-relevant features by generating channel-wise weights based on local and global information. It computes:
\[
g = \sigma(W_1 \cdot \text{GAP}(x + x_{\text{global}})),
\]
where \( x_{\text{global}} \) is a global feature map, \(\text{GAP}\) is Global Average Pooling, and \(\sigma\) is the sigmoid activation function. The combined input \(x + x_{\text{global}}\) captures both local and global features, ensuring the network focuses on critical semantic regions. The weights \( g \) reweight the input features as:
\[
x_{\text{att}} = g \odot x.
\]

\subsubsection{Deep Supervision}
\label{sec:Deep Supervision}
\textbf{Deep Supervision} \cite{lee2014deeplysupervisednets,li2022comprehensivereviewdeepsupervision} aligns feature learning with classification objectives by optimizing the equivalent matrix \( W_{\text{equiv}} \). It minimizes a task-specific loss function, guiding feature transformations to capture both local details and global semantics:
\[
\mathcal{L}_{\text{pred}} = \mathcal{L}\big(W_2 W_1 \text{ReLU}(x + x_{\text{global}})\big),
\]
where \( W_1 \) and \( W_2 \) are weight matrices, \( x \) represents the feature map, and \( x_{\text{global}} \) is a global context feature map. The ReLU activation ensures that only positive feature contributions are considered.

Deep supervision refines shallow-layer representations, enabling the model to capture finer details. During testing, only the final layer is used for predictions, improving computational efficiency as the intermediate deep-supervision components are removed.

The loss during pretraining and fine-tuning is defined as:
\[
\text{Loss} =
\begin{cases}
\mathcal{L}\left(\sum \hat{y}\right), & \text{for pretraining}, \\
\sum \mathcal{L}(\hat{y}), & \text{for fine-tuning}.
\end{cases}
\]
Pretraining optimizes over the overall task, while fine-tuning provides layer-specific supervision to capture task-relevant features at different scales.

\subsubsection{Equivalent Matrix \& Saliency Score}
The \textbf{Equivalent Matrix} \( W_{\text{equiv}} = W_2 W_1 \) models interactions between local and global features. The Attention Gate ensures fine-grained channel selection, while Deep Supervision aligns feature representations globally. Together, they enable \( W_{\text{equiv}} \) to capture high-resolution texture details and task-relevant semantics crucial for various applications, including medical imaging and natural language processing.

In Convolutional Neural Networks (CNNs), the Equivalent Matrix facilitates the generation of Class Activation Maps (CAMs), which highlight the regions of the input image that are most influential for a specific class prediction. For each layer \( l \), the CAM is computed as:
\[
\text{CAM}^{(l)}(x, y) = \sum_{k=1}^{C} w_{\text{adjusted}, k}^{(l)} \cdot f_{k}^{(l)}(x, y),
\]
\begin{itemize}[left=0pt]
    \item \( w_{\text{adjusted}, k}^{(l)} \) are the adjusted weights derived using Non-Negativity and Salience Sharpness (Section~\ref{Neuron Analysis}).
    \item \( f_{k}^{(l)}(x, y) \) are spatial activations of channel \( k \) at layer \( l \)
\end{itemize}
The final CAM is obtained by aggregating CAMs across all layers with a weighting factor \( \alpha^{l} \):
\[
\text{CAM} = \sum_{l} \alpha^{l} \text{CAM}^{(l)},
\]
where \( \alpha^{l} \) is typically set to \( 0.9 \) to balance the contributions from shallow-layer fine details and deep-layer global semantics.

In Transformer-based architectures, such as BERT, saliency is represented through saliency scores rather than saliency maps due to the sequential nature of the data. The saliency score for each token is computed by aggregating contributions from multiple layers. To address the over-smoothing problem, we set the number of layers \( L \) to 3, as activations tend to become too smooth after the third layer in certain datasets, such as AG News. This empirical choice prevents over-smoothing from diminishing the quality of saliency explanations. The process involves the following steps:
\[
S^{(l,c)}(j) = \sum_{k=1}^{D} W_{\text{equiv},c,k} \cdot A^{(l)}(j,k),
\]
\[
S^{(c)}(j) = \sum_{l=1}^{L} \alpha^{l} S^{(l,c)}(j),
\]
\begin{itemize}[left=0pt]
    \item \( S^{(l,c)}(j) \) is the saliency score of the \( j \)-th token at layer \( l \) for class \( c \).
    \item \( W_{\text{equiv},c,k} \) are the weights from the Equivalent Matrix corresponding to class \( c \) and feature \( k \).
    \item \( A^{(l)}(j,k) \) is the activation value of the \( j \)-th token and \( k \)-th feature at layer \( l \).
\end{itemize}

The final saliency score \( S^{(c)}(j) \) quantifies the importance of token \( j \) for class \( c \), capturing contributions from all layers weighted appropriately. By limiting the number of layers \( L \), we ensure that early layers can capture rich details while mitigating the potential for over-smoothing in deeper layers.

\subsection{Integration}

In this section, we describe how MHEX is integrated into residual networks, such as ResNet, and Transformer models, like BERT, to enhance feature representation and interpretability while maintaining the original architecture.

As illustrated in Figure~\ref{fig:den_resnet}, MHEX is deployed between the residual blocks of ResNet.
\begin{figure}[h!]
    \centering
    \includegraphics[width=1\linewidth]{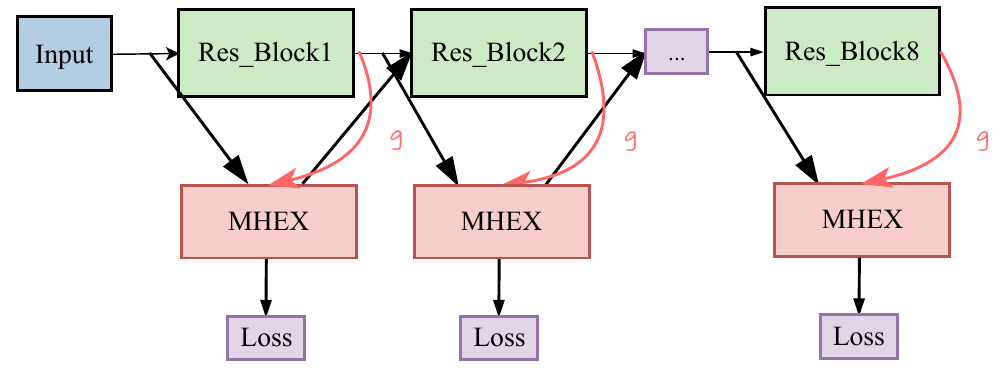}
    \caption{The internal structure of the Multi-Head Explainer (MHEX) integrated into a residual network (ResNet).}
    \label{fig:den_resnet}
\end{figure}

\begin{figure}[ht!]
    \centering
    \includegraphics[width=1\linewidth]{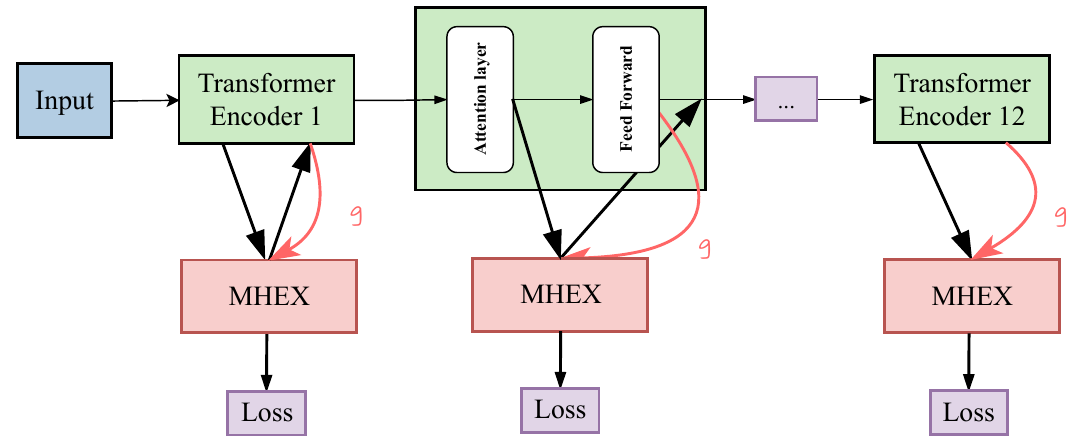}
    \caption{Architecture of MHEX integrated into a Transformer model (e.g., BERT).}
    \label{fig:den_transformer}
\end{figure}

Similarly, for Transformer models such as BERT, we insert MHEX between the attention layers and feed-forward layers, as shown in Figure~\ref{fig:den_transformer}. This integration allows us to extract detailed saliency scores, enhancing the model's interpretability. Specifically, after the attention mechanism, we apply Layer Normalization to stabilize the activations:
\[
x_{\text{att}} = \text{LayerNorm}(x_{\text{att}}),
\]
where \( x_{\text{att}} \) refers to the output of the attention mechanism, as previously defined in Section~\ref{attentiongate}.

\subsection{Local \& Global Analysis}
\label{Neuron Analysis}
In the early layers of the model, saliency scores often exhibit uncertainty due to limited semantic information, leading the model to mistakenly highlight background or noise as important activations. To mitigate this, we impose a \textbf{non-negativity constraint} by applying the ReLU activation function, which projects feature representations onto a non-negative subspace. This constraint effectively removes irrelevant features and reduces entropy (see Appendix~\ref{Entropy Reduction}), thereby facilitating more accurate neuron analysis and enhancing the interpretability of the equivalent matrix \( W_{\text{equiv}} \).

At a local level, focusing on individual neurons, we leverage the non-negativity constraint alongside a parameter \( \alpha \) (Appendix~\ref{Non-Negativity}) to decompose the weights in \( W_{\text{equiv}} \) into positive and negative components. By controlling the influence of negative contributions through \( \alpha \), we ensure that saliency scores emphasize features that positively contribute to class predictions. 

Globally, considering all neurons collectively, we introduce \textbf{salience sharpness} (\( \text{SS} \)) (see Appendix~\ref{Salience Sharpness}), which quantifies the specificity of each feature's contribution to class predictions across the entire network. By analyzing the distribution of salience scores in both positive and negative domains, \( \text{SS} \) highlights the most relevant features for the target class, ensuring that the saliency maps accurately reflect the integrated contributions of all neurons.


\subsection{Quantitative Study}

To quantitatively evaluate the quality of saliency scores generated by MHEX, we utilize several metrics that assess the impact of salient regions on model predictions.

\subsubsection{Average Drop (AVG Drop)}

The \textbf{Average Drop (AVG Drop)} metric measures the decrease in the model's confidence after the removal of salient regions, reflecting their importance in decision-making~\cite{chattopadhay2018grad}. It is computed as:
\[
\text{AVG Drop} = \frac{1}{N} \sum_{i=1}^N \max\left(0, \frac{p_{\text{orig}}^i - p_{\text{mask}}^i}{p_{\text{orig}}^i}\right),
\]
where \( p_{\text{orig}}^i \) and \( p_{\text{mask}}^i \) are the model's prediction confidences on the original and masked inputs for the \( i \)-th sample, respectively.

\subsubsection{Soft Average Drop (SAD)}

Soft Average Drop (SAD) improves upon AVG Drop by employing a soft replacement strategy that preserves the overall structure of the input, focusing the evaluation on saliency map quality. Due to the typically homogeneous image distribution in medical imaging, and to avoid the bias introduced by "hard" removal and confounders caused by model generalization issues, we propose this approach:
\[
I_{\text{soft}}(x, y) = I(x, y) \cdot (1 - \text{CAM}(x, y)) + \mu \cdot \text{CAM}(x, y),
\]
where \( I(x, y) \) is the original input intensity at location \( (x, y) \), and \( \mu \) is the mean intensity of the input.

\subsubsection{Effective Average Drop (EAD)}

We introduce the \textbf{Effective Average Drop (EAD)}, which incorporates the saliency map's area into the evaluation using an area-based weighting function \( f(x) \) (Appendix~\ref{appendix: EAD Weighting Function}). This function penalizes saliency maps that are excessively sparse or overly distributed, promoting compact and informative explanations. The EAD metric is computed as:
\[
\text{EAD} = \frac{1}{N} \sum_{i=1}^N \left( \text{Drop}_i \cdot f(x_i) \right),
\]
\( \text{Drop}_i \) is the confidence drop for the \( i \)-th sample, and \( x_i \) is the proportion of the saliency map’s area for that sample.


\subsection{Collaboration Analysis}
In the MHEX framework, the Attention Gate (AG) and Deep Supervision (DS) modules have distinct optimization objectives. To investigate their collaboration, we analyze the direction of gradients during training.

Specifically, we compute the cosine similarity between the gradients of the Equivalent Matrix \( W_{\text{equiv}} \) with respect to \( W_1 \), contributed by the Attention Gate (\( \nabla_{\text{AG}}^{W_1} \)) and Deep Supervision (\( \nabla_{\text{DS}}^{W_1} \)). These gradients are defined as:
\begin{itemize}[left=0pt]
    \item \( \nabla_{\text{AG}}^{W_1} = \frac{\partial \mathcal{L}_{\text{DS}}^{(l+1)}}{\partial W_1} \), where \( \mathcal{L}_{\text{DS}}^{(l+1)} \) is the loss from the Deep Supervision module in the next layer (\( l+1 \)).
    \item \( \nabla_{\text{DS}}^{W_1} = \frac{\partial \mathcal{L}_{\text{DS}}^{(l)}}{\partial W_1} \), where \( \mathcal{L}_{\text{DS}}^{(l)} \) is the loss from the current layer (\( l \)).
\end{itemize}

The cosine similarity between gradients is calculated as:
\[
\text{Cosine Similarity} = \frac{\nabla_{\text{AG}}^{W_1} \cdot \nabla_{\text{DS}}^{W_1}}{\|\nabla_{\text{AG}}^{W_1}\| \|\nabla_{\text{DS}}^{W_1}\| + \epsilon},
\]
where \( \epsilon \) is a small constant to prevent division by zero.

In the Experiments section, we apply this metric to assess the collaboration strength between AG and DS across multiple layers and evaluate its impact on the quality of the saliency scores.

\section{Experiments}
\subsection{Dataset}

We evaluate our proposed method on three datasets: \textbf{ImageNet1k}, \textbf{MedMNIST}, and \textbf{AG News}. These datasets provide diverse challenges across general-purpose and biomedical image classification tasks.

\textbf{ImageNet1k} is a large-scale dataset widely used for benchmarking deep learning models~\cite{deng2009imagenet}. It contains 1,000 object categories with approximately 1.28 million training images and 50,000 validation images.

\textbf{MedMNIST} is a collection of biomedical image datasets designed for lightweight classification tasks~\cite{medmnistv1, medmnistv2}. It includes both 2D and 3D datasets across various data modalities and task types.

\textbf{AG News}\footnote{\url{http://www.di.unipi.it/~gulli/AG_corpus_of_news_articles.html}} is a widely recognized benchmark for text classification tasks, comprising 120,000 training samples and 7,600 test samples across four distinct classes: World, Sports, Business, and Sci/Tech. The dataset consists of news articles, providing a diverse range of topics and writing styles.

Details of training and fine-tuning configurations, can be found in Appendix~\ref{appendix:Training and Fine-Tuning Details}.

\subsection{Results for CNNs}

Table \ref{tab:training_results} summarizes the performance of \textbf{MHEX-Net} compared to baseline ResNet-18 models. All baseline metrics are derived from the official MedMNIST and PyTorch benchmarks for ResNet-18~\cite{medmnistv2}. \textbf{MHEX-Net} consistently outperforms the baseline in most tasks, achieving higher accuracy with minimal additional parameters. Specifically, for ResNet-18, the added parameters amount to \(1920 \times \text{n}_{\text{classes}} + 0.69M \).

\begin{table}[h!]
    \centering
    \resizebox{\columnwidth}{!}{%
        \begin{tabular}{lcc}
            \toprule 
            \textbf{Task} & \textbf{MHEX-Net ACC (\%)} & \textbf{Benchmark ACC (\%)} \\
            \midrule 
            \textbf{ImageNet1k} & $\textbf{70.57} \pm 0.31 $ & 69.75 \\ 
            \textbf{PathMNIST} & \textbf{$\textbf{95.18} \pm 1.09 $} & 90.90 \\ 
            \textbf{OrganAMNIST} & \textbf{ $\textbf{97.66} \pm 1.17 $ } & 95.10 \\ 
            \textbf{TissueMNIST} &  $66.81 \pm 0.67 $ & \textbf{68.10} \\ 
            \textbf{BloodMNIST} & $\textbf{96.66} \pm 0.48 $ & 96.30 \\
            \midrule 
            \textbf{AG News*} & $93.63 \pm 0.20$ & $\textbf{94.56} \pm 0.37$ \\ 
            \bottomrule 
        \end{tabular}%
    }
    \caption{Comparison of accuracy between \textbf{MHEX-Net} and baseline ResNet-18 benchmarks. Benchmark ACC values are sourced from the official reports.\protect\footnotemark[2] \protect\footnotemark[3] \\ * Comparison between \textbf{MHEX-BERT} and BERT, results are based on experimental data.}
    \label{tab:training_results}
\end{table}

\footnotetext[2]{MedMNIST Benchmarking Report: \href{https://medmnist.com/\#Benchmarking}{MedMNIST}}
\footnotetext[3]{Torchvision Documentation: \href{https://pytorch.org/vision/stable/models/generated/torchvision.models.resnet18.html}{Torchvision ResNet-18 Documentation}}

\subsection{Saliency Score Analysis}

To assess the interpretability and localization capabilities of the \textbf{MHEX-Net} framework, we compared saliency scores generated by \textbf{MHEX-Net}, Grad-CAM, SHAP (with blur masking using 20k $\times$ 256 or 50k $\times$ 256 evaluations), and Layer-CAM across five representative tissue types.

\begin{figure*}[!ht]
    \centering
    \setlength{\arrayrulewidth}{0pt}     
    \begin{tabular}{ccccc}
      Original & \textbf{MHEX} & Grad-CAM & SHAP & Layer-CAM \\
        \includegraphics[width=0.14\textwidth]{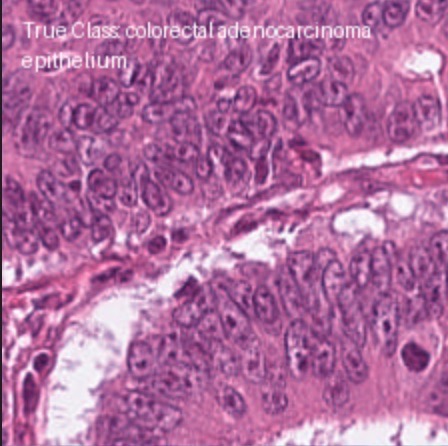} &
        \includegraphics[width=0.14\textwidth]{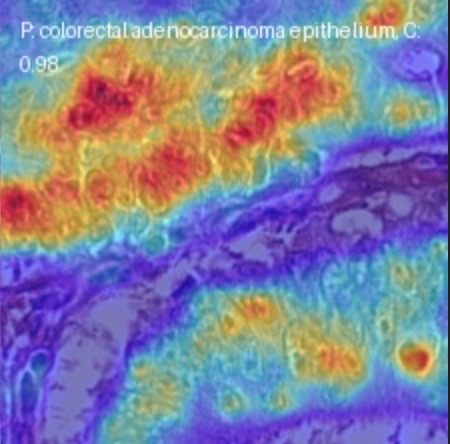} &
        \includegraphics[width=0.14\textwidth]{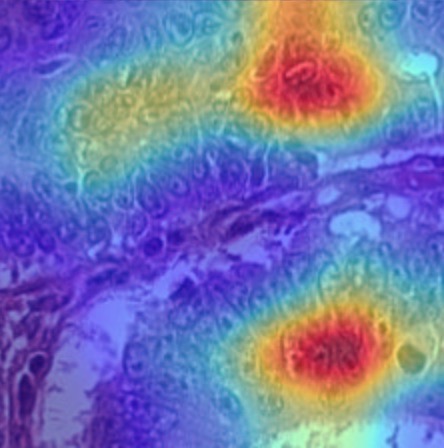} &
        \includegraphics[width=0.14\textwidth]{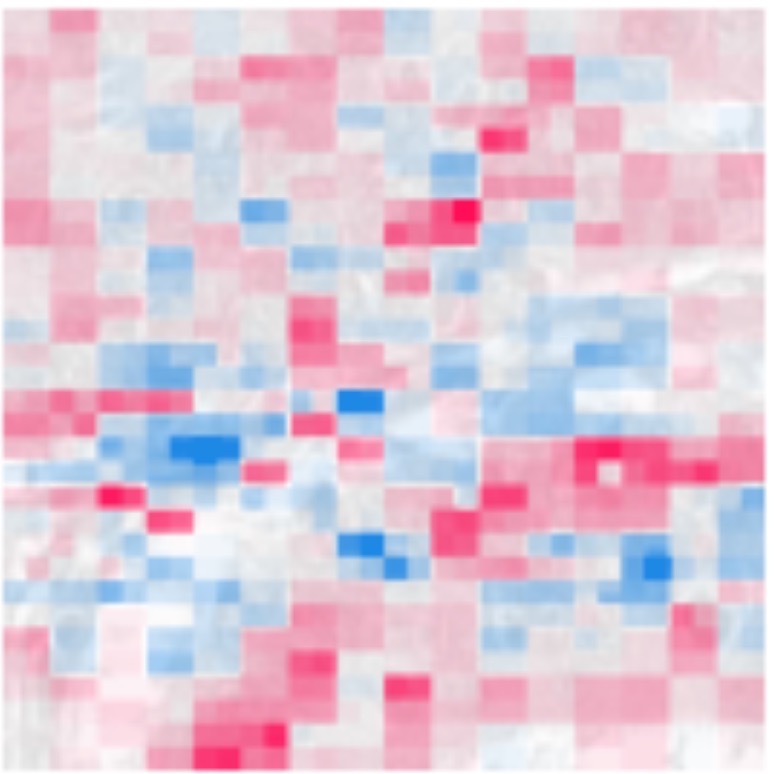} &
        \includegraphics[width=0.14\textwidth]{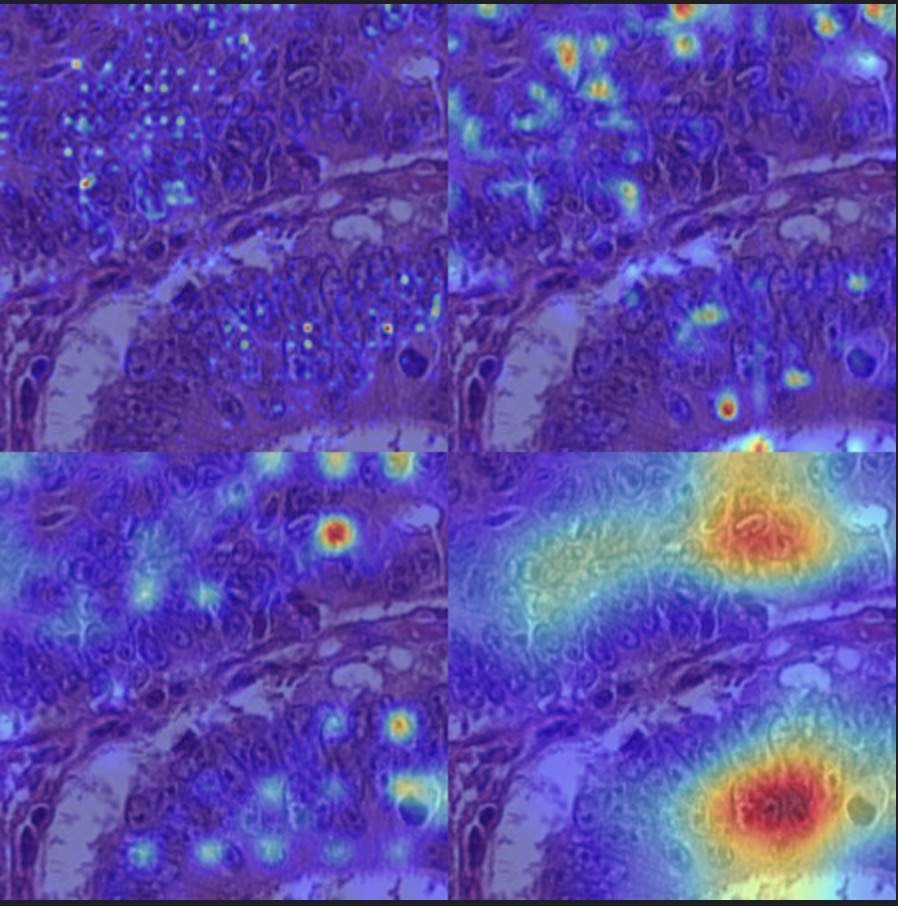} \\

        \includegraphics[width=0.14\textwidth]{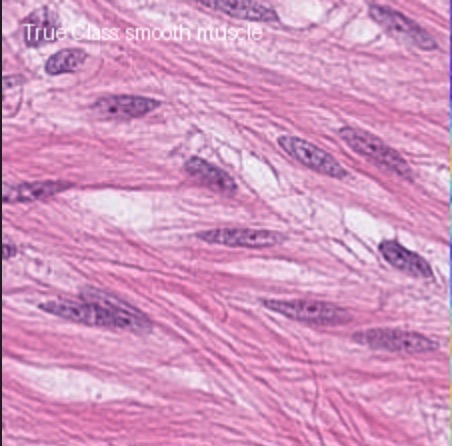} &
        \includegraphics[width=0.14\textwidth]{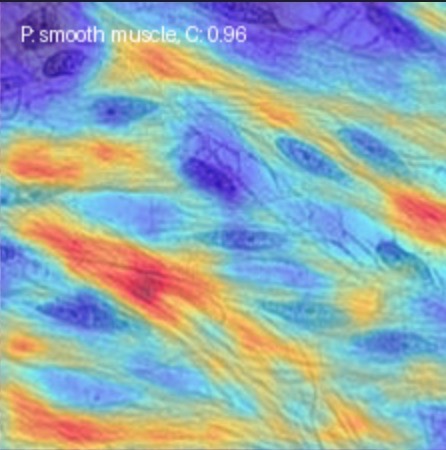} &
        \includegraphics[width=0.14\textwidth]{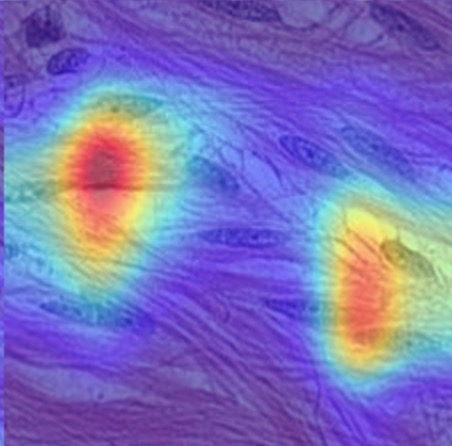} &
        \includegraphics[width=0.14\textwidth]{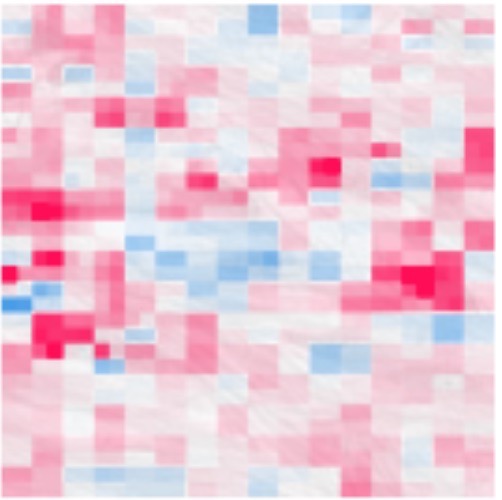} &
        \includegraphics[width=0.14\textwidth]{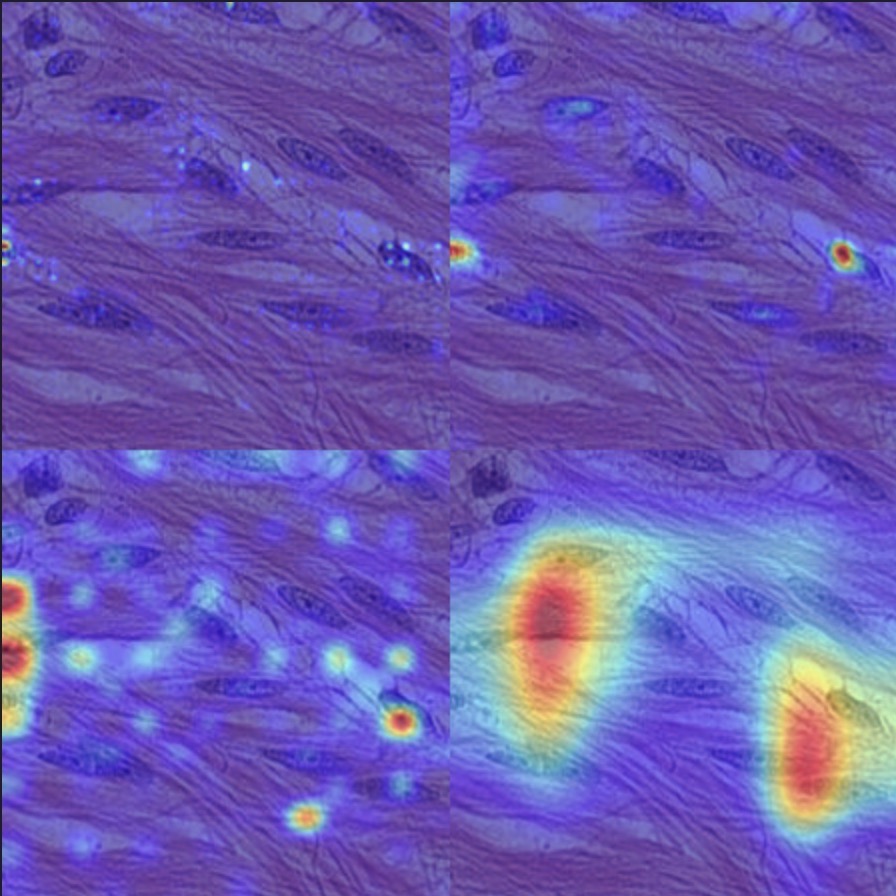} 
      
    \end{tabular}
    \caption{
        Saliency score comparison for smooth muscle tissue and colorectal adenocarcinoma epithelium. Colorectal adenocarcinoma epithelium is characterized by irregularly arranged epithelial cells with enlarged nuclei and poor glandular differentiation. Smooth muscle tissue consists of elongated fibers arranged in parallel, with nuclei distributed along the fibers. \textbf{MHEX} effectively highlights the fibers and cancerous regions with uniform coverage, closely resembling semantic segmentation.
    }
    \label{tab:combined}
\end{figure*}

\begin{figure*}[h!]
    \centering
    \setlength{\arrayrulewidth}{0pt}
    \setlength{\tabcolsep}{1pt} 
    \renewcommand{\arraystretch}{1} 
    \begin{tabular}{|>{\centering\arraybackslash}m{0.03\textwidth}|c|c|c|c|c|c|c|c|}
        \hline
         & Original Image & \textbf{MHEX} & Grad-CAM &  & Original Image &  \textbf{MHEX}  & Grad-CAM\\
        \hline

        \rotatebox[origin=c]{90}{\textbf{Limpkin}} &
        \raisebox{-0.5\height}{\includegraphics[width=0.13\textwidth]{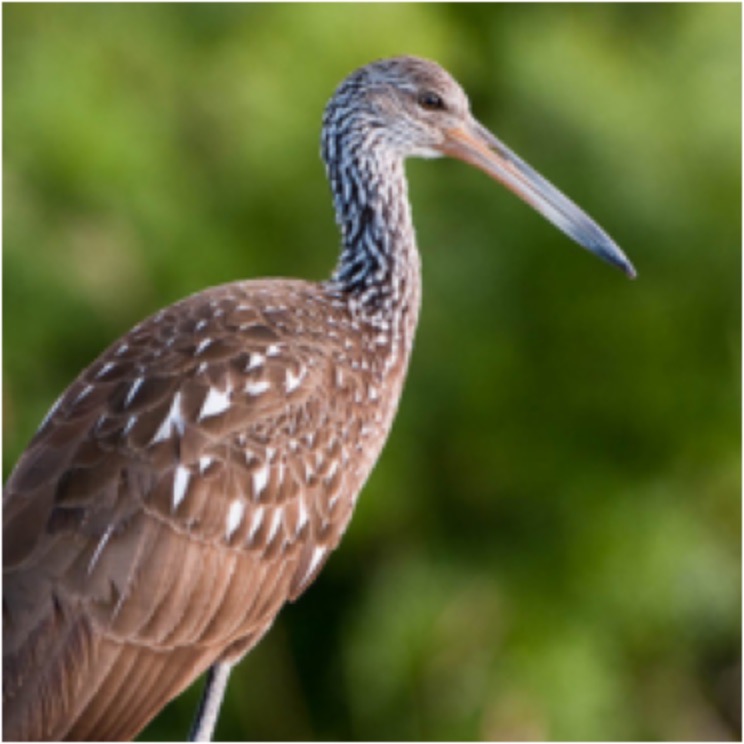}} &
        \raisebox{-0.5\height}{\includegraphics[width=0.13\textwidth]{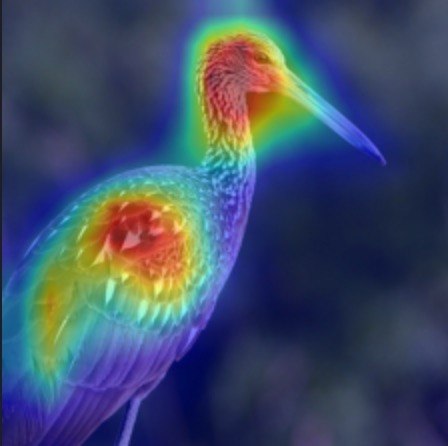}} &
        \raisebox{-0.5\height}{\includegraphics[width=0.13\textwidth]{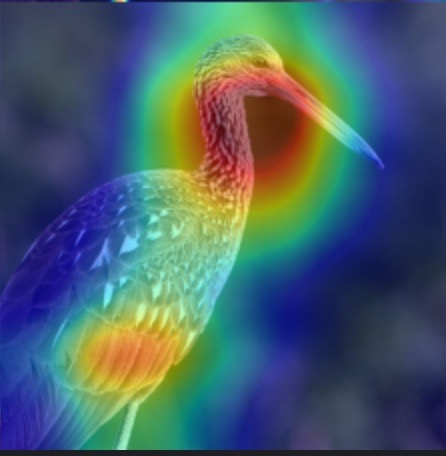}} &

       \rotatebox[origin=c]{90}{  \textbf{Sandpiper}} &
        \raisebox{-0.5\height}{\includegraphics[width=0.13\textwidth]{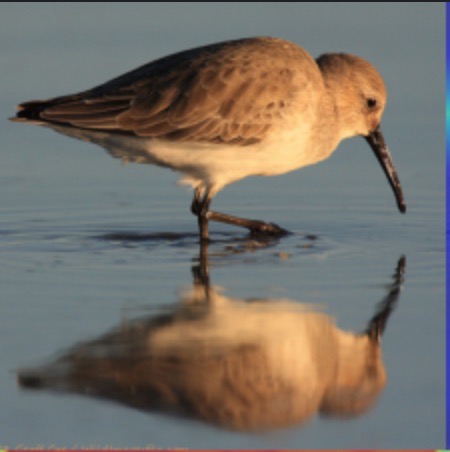}} &
        \raisebox{-0.5\height}{\includegraphics[width=0.13\textwidth]{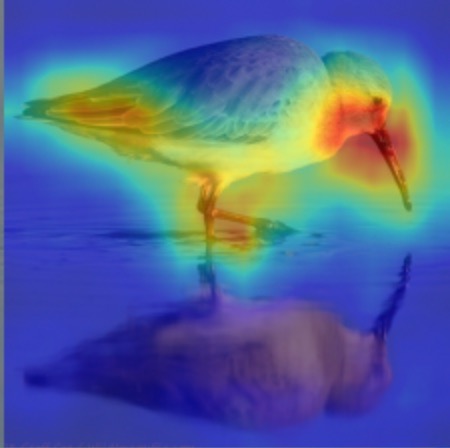}} &
        \raisebox{-0.5\height}{\includegraphics[width=0.13\textwidth]{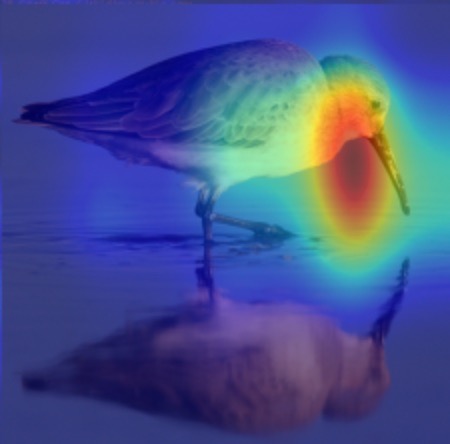}} \\
        \hline
        
    \end{tabular}
    \caption{Comparison of original images, MHEX, and Grad-CAM saliency maps on selected ImageNet1k classes.}
    \label{tab:imagenet1k_maintext}
\end{figure*}

\textbf{Image Classification}: \textbf{MHEX-Net} consistently demonstrated superior coverage, closely resembling semantic segmentation, while other methods exhibited characteristic strengths and limitations, such as focusing on localized high-level features or producing fragmented scores.The results are illustrated in Figure~\ref{tab:combined} (colorectal adenocarcinoma epithelium and smooth muscle tissue) and Figure~\ref{tab:imagenet1k_maintext} (ImageNet).

For \textbf{Text Classification} tasks on the AG News dataset, \textbf{MHEX-Net} effectively captured important tokens relevant to class predictions. Table~\ref{tab:salience_transformer} presents saliency scores for selected test samples from the Business, Sci/Tech, and Sports categories, illustrating how each method assigns importance to different tokens. \textbf{MHEX-Net} consistently reveals rich and detailed insights in this task.

\begin{table*}[!h]
    \centering
    \setlength{\arrayrulewidth}{0pt}
    \renewcommand{\arraystretch}{1.4} 
    \includegraphics[width=1\textwidth]{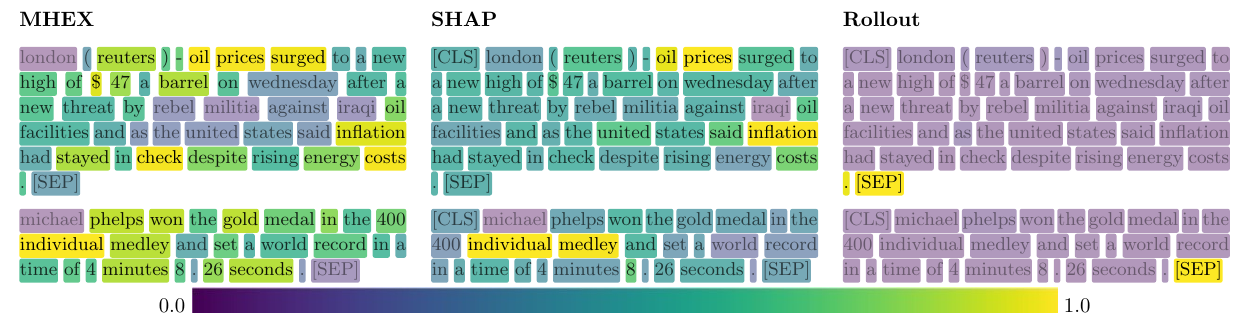}
    \caption{Saliency Scores Comparison for Selected AG News Samples. Each column compares the saliency scores generated by \textbf{MHEX}, SHAP, and Attention Rollout. The color intensity represents the importance score.}
    \label{tab:salience_transformer}
\end{table*}
\textit{Additional results are available in Appendix~\ref{appendix:C} for further reference.}

\subsection{Local \& Global Analysis}
In this study, we investigate the impact of two critical hyperparameters, $\alpha$ and $SS$ (Salience Sharpness), on the quality and interpretability of the saliency maps generated by our \textbf{MHEX-Net} framework.

\begin{figure}[!h]
    \centering
    \setlength{\arrayrulewidth}{0pt}
    \setlength{\tabcolsep}{1pt}
    \begin{tabular}{ccc|ccc}
        \hline
         $\alpha = 1$ & $\alpha = 0.5$ & $\alpha = 0$ & $SS = 0$ & $SS = 0.3$ & $SS = 0.5$ \\
        \hline
            \includegraphics[width=0.077\textwidth]{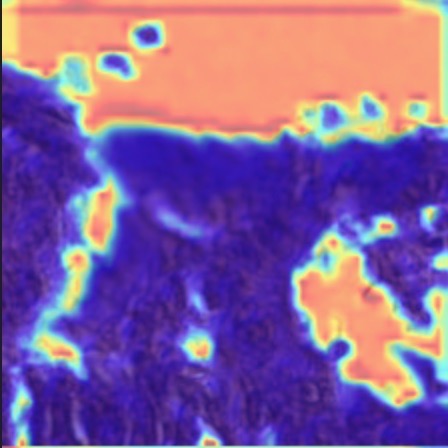} & 
            \includegraphics[width=0.077\textwidth]{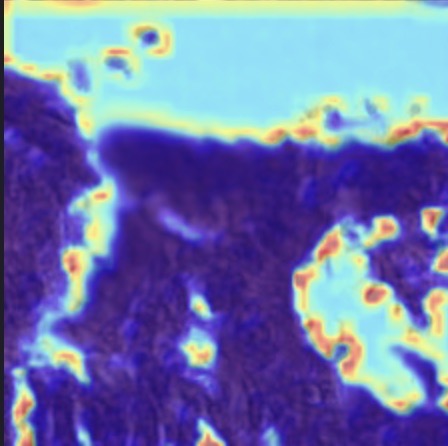} & 
            \includegraphics[width=0.077\textwidth]{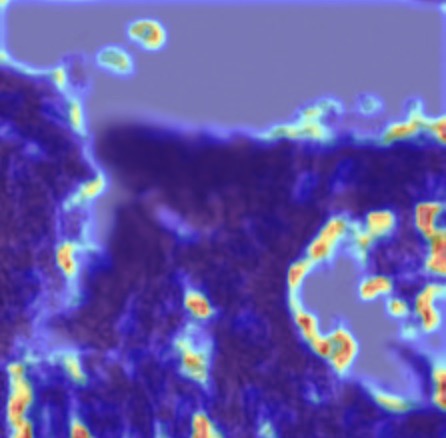} &
            \includegraphics[width=0.077\textwidth]{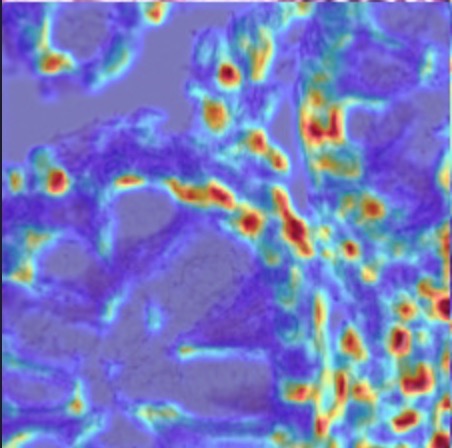} & 
            \includegraphics[width=0.077\textwidth]{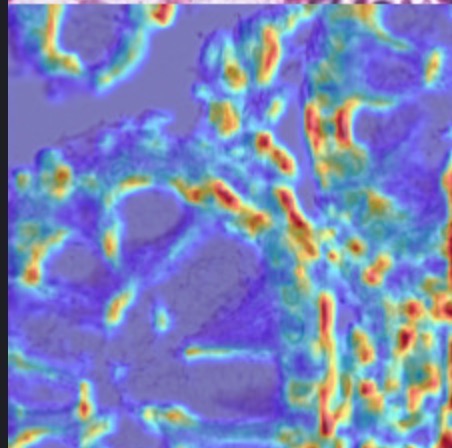} & 
            \includegraphics[width=0.077\textwidth]{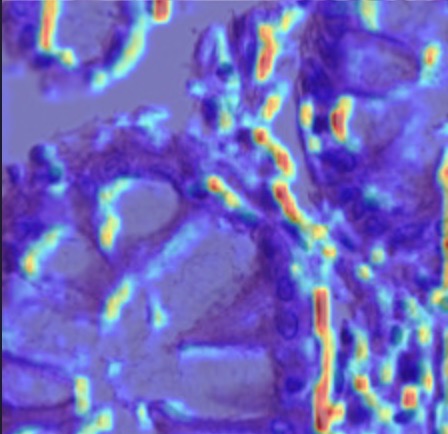} \\
         
            \includegraphics[width=0.077\textwidth]{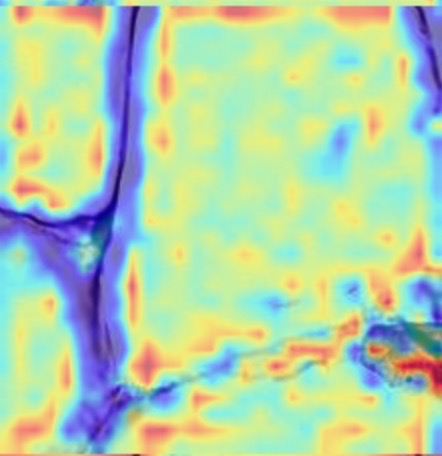} & 
            \includegraphics[width=0.077\textwidth]{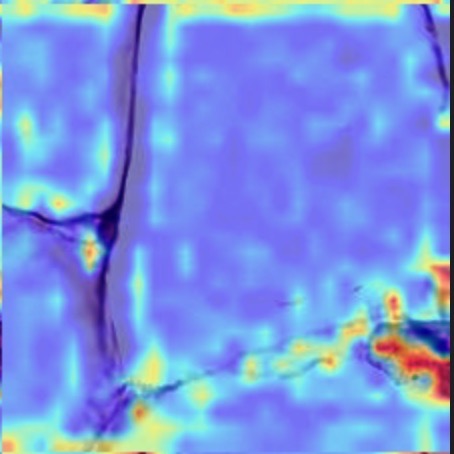} & 
            \includegraphics[width=0.077\textwidth]{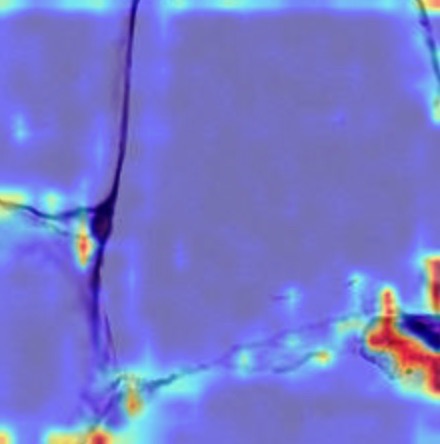} &
            \includegraphics[width=0.077\textwidth]{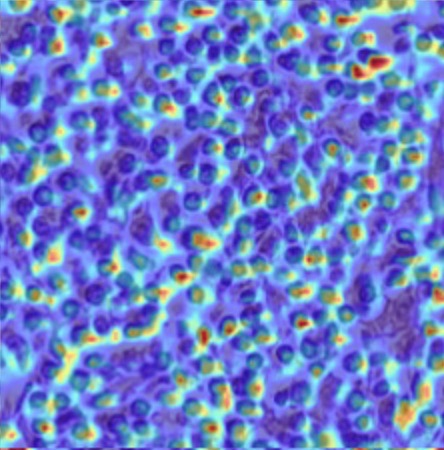} & 
            \includegraphics[width=0.077\textwidth]{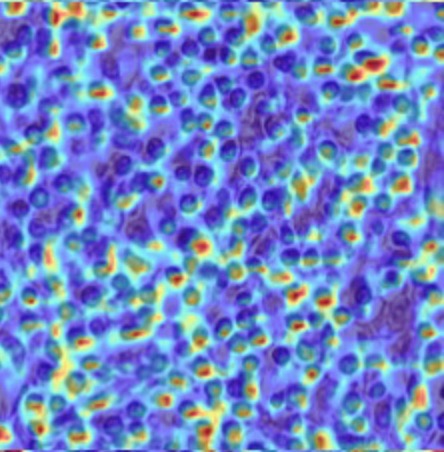} & 
            \includegraphics[width=0.077\textwidth]{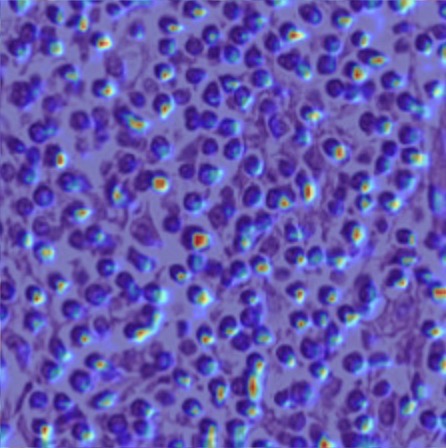} \\
    \end{tabular}
    \caption{Saliency maps of early layers with varying $\alpha$ and $SS$ values. Decreasing $\alpha$ filters out background noise, while higher $SS$ values produce sharper and more focused saliency maps.}
    \label{tab:combined_hyperparameters}
\end{figure}

The hyperparameter $\alpha$ plays a crucial role in enhancing the clarity of early-layer interpretations by filtering out noisy channels. By tuning $\alpha$, we effectively exclude channels with negative contributions, resulting in sharper and more interpretable saliency maps. As illustrated in Figure~\ref{tab:combined_hyperparameters}, decreasing $\alpha$ significantly reduces early-layer uncertainties. For instance, when $\alpha = 0$, only channels with positive contributions are included in the calculation of the saliency map.

With $\alpha$ fixed at 0, we further explore the impact of $SS$, another hyperparameter designed to filter uncertain channels by prioritizing those that exhibit strong connections to specific class neurons. Higher $SS$ values correspond to sharper and more focused saliency maps, as demonstrated in Figure~\ref{tab:combined_hyperparameters}. Increasing $SS$ enhances the concentration of saliency in regions highly relevant to the class predictions, thereby improving the interpretability of the model's decision-making process.

Based on our observations, we recommend setting $SS = \frac{1}{n_\text{class}} + \epsilon$, where $\epsilon \in [0.1, 0.3]$. This configuration effectively balances the sharpness and coverage of saliency maps across different classes. Additionally, for different datasets, $\alpha$ values in the range $[0, 0.5]$ generally yield favorable results.

\subsection{Quantitative Study}

We evaluate the performance of our \textbf{MHEX-Net} and \textbf{MHEX-BERT} models in comparison to Grad-CAM, Layer-CAM, and SHAP using the metrics \textbf{Average Drop (AVG Drop)}, \textbf{Soft Average Drop (SAD)}, and \textbf{Effective Average Drop (EAD)} on the PathMNIST, BloodMNIST, and AG News datasets. 

\begin{table*}[!t]
    \centering
    \resizebox{\textwidth}{!}{%
       \begin{tabular}{l|ccc|ccc|ccc}
            \toprule
            \textbf{Task} & \multicolumn{3}{c|}{\textbf{Soft Average Drop (SAD)}} & \multicolumn{3}{c|}{\textbf{Average Drop (AVG Drop)}} & \multicolumn{3}{c}{\textbf{Effective Average Drop (EAD)}} \\
            \cline{2-10}
            & MHEX & Grad-CAM & Layer-CAM & MHEX & Grad-CAM & Layer-CAM & MHEX & Grad-CAM & Layer-CAM \\
            \midrule
            PathMNIST & \textbf{0.2082} & 0.1181 & 0.1069 & \textbf{0.4276} & 0.1522 & 0.1247 & \textbf{0.0710} & 0.0650 & 0.0600 \\
            BloodMNIST & 0.5007 & 0.4993 & \textbf{0.5078} & 0.7212 & 0.7765 & \textbf{0.7837} & \textbf{0.1894} & 0.1106 & 0.0808 \\
        \midrule
           \multicolumn{1}{l}{AG News*}  & \multicolumn{3}{c|}{} & \multicolumn{3}{l}{ MHEX:0.1245 \hfill SHAP: \textbf{0.2657} } & \multicolumn{3}{|c}{} \\
            \bottomrule

        \end{tabular}
    }
    \caption{
        Comparison of Average Drop (AVG Drop), Soft Average Drop (SAD), and Effective Average Drop (EAD) metrics on PathMNIST and BloodMNIST datasets for \textbf{MHEX-Net}, Grad-CAM, and Layer-CAM methods. *For AG News, \textbf{MHEX-BERT} and SHAP are compared using the AVG Drop metric.
    }
    \label{tab:quantitative_results}
\end{table*}

The results are summarized in Table~\ref{tab:quantitative_results}. Bold values indicate the best performance in each category.

In the \textbf{PathMNIST} dataset, which contains abundant fine-grained features, Grad-CAM and Layer-CAM often produce suboptimal results, struggling to capture intricate patterns. \textbf{MHEX-Net} effectively captures these details, resulting in superior performance across all three metrics.

For the \textbf{BloodMNIST} dataset, the situation differs. Since BloodMNIST images feature distinct cell bodies without pervasive fine details, explanation methods like Grad-CAM and Layer-CAM can effectively highlight the main structures, yielding reasonable results. This is showed in their comparable SAD and AVG Drop metrics. However, when considering the \textbf{EAD} metric, which accounts for the effectiveness of the explanation relative to the area of the saliency map, MHEX outperforms the other methods significantly.

In the \textbf{AG News} dataset, we compared \textbf{MHEX-BERT} against SHAP using the AVG Drop metric. To evaluate the impact of saliency methods on the BERT model’s performance, we selected the top 10\% of saliency tokens for perturbation using the [MASK] token across the first 512 samples of the AG News dataset. This threshold balances computational efficiency, addressing the high cost associated with SHAP.

The results showed that \textbf{SHAP} achieved a higher AVG Drop, indicating that it is more effective in identifying critical tokens that influence the model's predictions. This suggests that while \textbf{MHEX-BERT} provides detailed saliency scores, SHAP’s theoretically grounded methodology performs better in pinpointing the most impactful tokens. 

One possible explanation is that SHAP’s scores are derived from drop-oriented methods based on Shapley values, which inherently consider each token's contribution across all possible feature subsets. In contrast, \textbf{MHEX-BERT}'s saliency scores are generated intrinsically from the model’s internal mechanisms without relying on feature perturbations, thereby avoiding the biases and limitations of drop-oriented approaches. 

Additionally, we found that using the [MASK] token for perturbation with both \textbf{MHEX-BERT} and SHAP did not significantly affect the model’s performance based on the AVG Drop metric. This contrasts with image tasks, where perturbing salient features typically leads to greater performance degradation. A possible explanation is that BERT has learned to effectively predict the [MASK] token during pretraining, enhancing its robustness to such perturbations. 

\textit{For Insertion and Deletion Curves, see: Appendix~\ref{appendix:Results of Insertion and Deletion Curves}}

\subsection{Collaboration Analysis \& Saliency Quality}

We analyze the interrelationships among collaboration strength (cosine similarity) across the last three MHEX-Blocks, model confidence \(p_{\text{orig}}\), and saliency map quality (SAD) on the PathMinist validation set, quantified using Pearson correlation coefficients (details in Appendix~\ref{appendix:pearson}). Figure~\ref{fig:correlation_triangle} illustrates these relationships.

\begin{figure}[htb]
    \centering
    \scalebox{0.7}{ 
\begin{tikzpicture}

    \coordinate (Block) at (1, 0);
    \coordinate (SAD) at (5, 0);
    \coordinate (p_orig) at (3, 3);

    \draw[thick] (Block) -- (SAD) -- (p_orig) -- cycle;

    \node[below] at (Block) {\textbf{MHEX 6 - 7}};
    \node[below] at (SAD) {\textbf{SAD}};
    \node[above] at (p_orig) {\textbf{$p_\text{orig}$}};

    \pgfmathanglebetweenpoints{\pgfpointanchor{Block}{center}}{\pgfpointanchor{SAD}{center}}
    \let\angleBlockSAD\pgfmathresult

    \pgfmathanglebetweenpoints{\pgfpointanchor{Block}{center}}{\pgfpointanchor{p_orig}{center}}
    \let\angleBlockPOrig\pgfmathresult

    \pgfmathanglebetweenpoints{\pgfpointanchor{SAD}{center}}{\pgfpointanchor{p_orig}{center}}
    \let\angleSADPOrig\pgfmathresult

    \def\offset{0.4} 

    \node[rotate=\angleBlockSAD] at ($(Block)!0.5!(SAD) + (0, \offset -0.6) $) {\scriptsize $0.1710$};
    \node[rotate=\angleBlockSAD] at ($(Block)!0.5!(SAD) + (0, \offset-0.9)$) {\scriptsize $0.0634$};
    \node[rotate=\angleBlockSAD] at ($(Block)!0.5!(SAD) + (0, \offset-1.2)$) {\scriptsize $0.1212$};

    \node[rotate=\angleBlockPOrig] at ($(Block)!0.5!(p_orig) + (-0.2, 0.2)$) {\scriptsize $-0.0277^*$};
    \node[rotate=\angleBlockPOrig] at ($(Block)!0.5!(p_orig) + (-0.5, 0.35)$) {\scriptsize $-0.1526$};
    \node[rotate=\angleBlockPOrig] at ($(Block)!0.5!(p_orig) + (-0.8, 0.5)$) {\scriptsize $-0.1299$};

    \node[rotate=\angleSADPOrig -180] at ($(SAD)!0.5!(p_orig) + (0.2, 0.2)$) {\scriptsize $-0.1765$};
\end{tikzpicture}}

    \caption{Pearson correlation coefficients across SAD, $p_\text{orig}$, and collaboration strengths. 
    All correlations are significant ($p = 0$), except for the value marked with $^*$, where $p = 0.0055$.}
    \label{fig:correlation_triangle}
\end{figure}
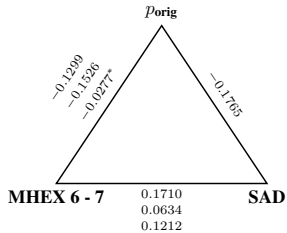

As the triangle shows, collaboration strength positively correlates with SAD, indicating that stronger collaboration improves the quality of saliency maps by aligning the highlighted regions with decision-critical features. 

At the same time, collaboration strength exhibits a negative correlation with model confidence (\( p_{\text{orig}} \)). One possible explanation is that the observed relationship arises due to an underlying causal link between SAD and \( p_{\text{orig}} \): when the model has higher confidence in a sample, it often produces a lower SAD. This statistical association might make it appear that collaboration strength reduces confidence.

\begin{figure}[htb]
    \centering
    \setlength{\tabcolsep}{2pt} 
    \renewcommand{\arraystretch}{1}
    \begin{tabular}{cccc}
    \includegraphics[width=0.11\textwidth]{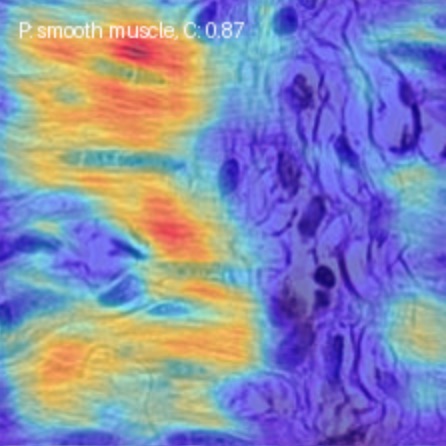}&
     \includegraphics[width=0.11\textwidth]{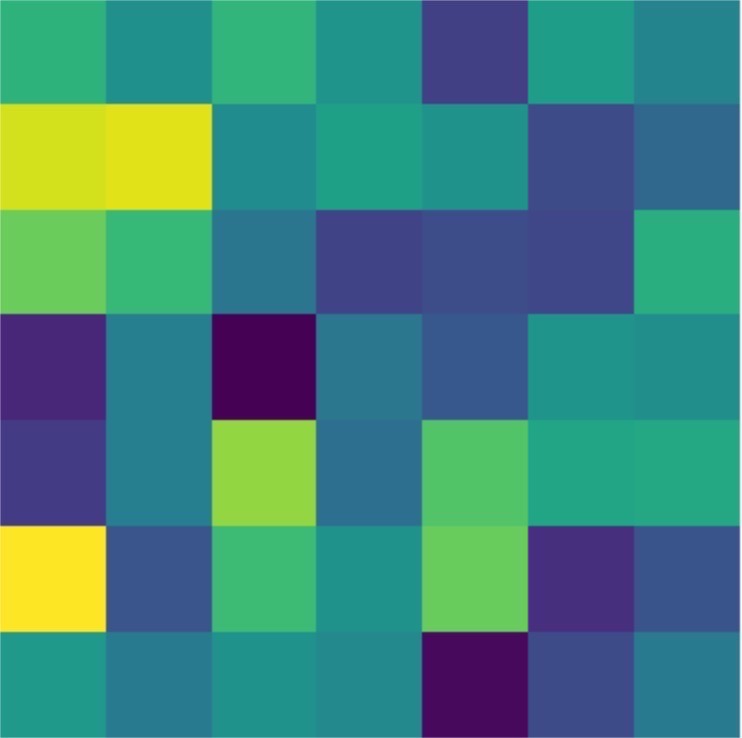} &
            \includegraphics[width=0.11\textwidth]{images_jpeg/colorectal_epithelium_DEN.jpeg} &
        \includegraphics[width=0.11\textwidth]{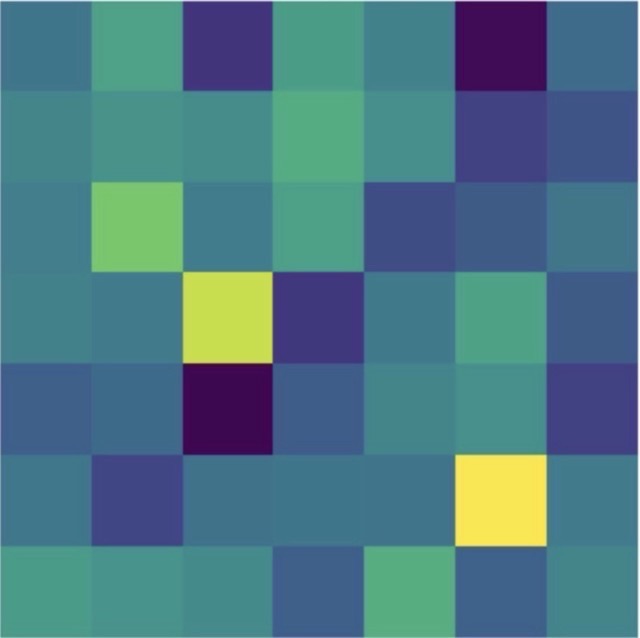}
         \\
        (a)&(b)&(c)&(d)\\
        \multicolumn{4}{c}{
            0.0  \includegraphics[width=0.4\textwidth]{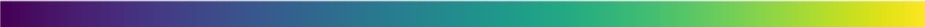} 1.0
        } \\

    \end{tabular}
    
    \caption{Block-wise quality analysis of saliency maps. The input image is divided into a \(7 \times 7\) grid, and gradient similarity is computed for each spatial block to represent MHEX's confidence in its explanations as heatmaps (b) and (c). Brighter blocks indicate higher confidence. Specifically, (b) shows MHEX's strong confidence in explaining the left half of the muscle regions, while (d) highlights a single cancer cell in the top-right corner with lower confidence despite its being marked in the salience map (c).}
    \label{tab:blockwise_quality}
\end{figure}

Based on this, we further evaluate saliency map quality at a finer spatial scale by dividing the input image into a \(7 \times 7\) grid and computing collaboration strength for each spatial block. This allows us to assess the localized quality of saliency maps. Examples are shown in Figure~\ref{tab:blockwise_quality}, where the collaboration metric serves as an internal measure of the model's confidence in aligning saliency maps with decision-critical features.

It is important to note that the proposed metric offers only one possible way to evaluate saliency map quality based on collaboration. Further investigations and validations are necessary to establish its robustness and generalizability, especially across more diverse datasets and tasks in real-world applications.

\section{Conclusion}
In this work, we introduced the \textbf{Multi-Head Explainer (MHEX)}, a comprehensive framework designed to enhance the explainability of both Convolutional Neural Networks (CNNs) and Transformer-based models. Our key contributions are summarized below:

\begin{itemize}[left=0pt, itemsep=0pt, topsep=0pt]
    \item \textbf{MHEX Framework}: Developed a versatile explanation framework compatible with CNNs and Transformers, with the potential for deployment in large language models (LLM) and AI agents, effectively addressing the limitations of existing methods in capturing fine-grained details and mitigating over-smoothing issues.
    
    \item \textbf{Neuron Analysis}: Presented a novel neuron analysis technique that enables more precise interpretations by examining individual neuron contributions, thereby enhancing the accuracy and reliability of saliency maps.
    
    \item \textbf{Collaboration Analysis}: Investigated the cooperative interactions between model components through gradient similarity measures, providing a new approach to assess the confidence and robustness of saliency scores.
\end{itemize}



\bibliographystyle{icml2025}
\bibliography{ref}

\appendix
\section{Mathematical Derivations}
\label{appendix:A}
\subsection{Entropy Reduction}
\label{Entropy Reduction}

To quantify the impact of non-negativity, we analyze entropy reduction through ReLU activation. Assume the output features \( T \) follow a standard normal distribution \( \mathcal{N}(0, 1) \), with differential entropy:
\[
H(T) = -\int_{-\infty}^{\infty} p(t) \ln p(t) \, dt = \frac{1}{2} \ln(2\pi e).
\]
After applying ReLU, the distribution becomes:
\[
p_{T_{\text{ReLU}}}(t) = 
\begin{cases} 
0.5 & \text{if } t = 0, \\
\frac{2}{\sqrt{2\pi}} e^{-t^2 / 2} & \text{if } t > 0.
\end{cases}
\]
The total entropy \( H(T_{\text{ReLU}}) \) consists of discrete and continuous components:
\[
H(T_{\text{ReLU}}) = H_{\text{discrete}} + H_{\text{continuous}},
\]
where:
\[
H_{\text{discrete}} = \frac{1}{2} \ln 2, \quad H_{\text{continuous}} = \frac{1}{2} \ln(2\pi e) - \ln 2.
\]
The overall entropy after ReLU activation is:
\[
H(T_{\text{ReLU}}) = \frac{1}{2} \ln(2\pi e) - \frac{1}{2} \ln 2.
\]
The entropy reduction can be expressed as:
\[
\Delta H = H(T) - H(T_{\text{ReLU}}) = \frac{1}{2} \ln 2 > 0.
\]

This reduction decreases information uncertainty, simplifies the system, and focuses on features with positive contributions. The resulting sparsity, with many neurons outputting zero, enhances both interpretability and analytical feasibility.

\subsection{Non-Negativity and Confidence Saliency Maps}
\label{Non-Negativity}

Building on the entropy reduction, we leverage the non-negativity constraint to generate "confidence saliency maps." ReLU's non-negativity ensures that a neuron’s contributions can be split into positive and negative components. For class \( i \), let \( w^i \in \mathbb{R}^C \) denote the weights connecting feature channels to the class. These are decomposed as:
\[
w_{\text{positive}}^{i} = \max(w^{i}, 0), \quad w_{\text{negative}}^{i} = \min(w^{i}, 0).
\]
To control the influence of negative contributions, we introduce a parameter \( \alpha \in [0, 1] \):
\[
w_{\text{adjusted}}^{i} = w_{\text{positive}}^{i} + \alpha \cdot w_{\text{negative}}^{i}.
\]
The Class Activation Map (CAM) for class \( i \) is then computed as:
\[
\text{CAM}^{i}(x, y) = \sum_{j=1}^{C} w_{\text{adjusted}}^{i}[j] \cdot f_{j}(x, y),
\]
where \( f_{j}(x, y) \) represents the activation of feature channel \( j \) at spatial location \( (x, y) \).

\subsection{Salience Sharpness}
\label{Salience Sharpness}

To further enhance interpretability, we analyze the “specificity” of each feature’s contribution to class predictions. In fully connected layers, a feature can contribute to the activations of neurons across multiple classes. To quantify this, we define salience sharpness (\( \text{SS} \)) for positive and negative contributions:
\begin{align*}
    \text{SS}_{\text{positive}}^{i}(j) &= 
    \frac{w_{\text{positive}}^{i}[j]}{\sum_{k=1}^{n} w_{\text{positive}}^{k}[j] + \epsilon}, \\
    \text{SS}_{\text{negative}}^{i}(j) &= 
    \frac{|w_{\text{negative}}^{i}[j]|}{\sum_{k=1}^{n} |w_{\text{negative}}^{k}[j]| + \epsilon}.
\end{align*}
Here, \( n \) is the total number of classes, and \( \epsilon > 0 \) is a small positive constant (e.g., \( 10^{-8} \)) to avoid division by zero.

The salience sharpness is used to adjust the weights, emphasizing class-specific contributions:
\[
w_{\text{final}}^{i}[j] = 
\begin{aligned}
    &\left( w_{\text{positive}}^{i}[j] \, \Big| \, \text{SS}{\text{positive}}^{i}(j) > ss  \right) \\ 
    + \alpha \cdot &\left( w_{\text{negative}}^{i}[j] \, \Big| \, \text{SS}{\text{negative}}^{i}(j) > ss \right).
\end{aligned}
\]

Finally, the CAM for class \( i \) is calculated using the adjusted weights:
\[
\text{CAM}^{i}(x, y) = \sum_{j=1}^{C} w_{\text{final}}^{i}[j] \cdot f_{j}(x, y).
\]
This refinement ensures that the CAM focuses on features with high specificity to the target class, improving both interpretability and robustness.

\subsection{Pearson Correlation Coefficient and p-value}
\label{appendix:pearson}

The Pearson correlation coefficient (\(r\)) measures the linear relationship between two variables \(X\) and \(Y\). It ranges from \(-1\) to \(1\):
\[
r = \frac{\sum_{i=1}^{n} (X_i - \bar{X})(Y_i - \bar{Y})}{\sqrt{\sum_{i=1}^{n} (X_i - \bar{X})^2}\sqrt{\sum_{i=1}^{n} (Y_i - \bar{Y})^2}},
\]
where \(X_i\) and \(Y_i\) are individual observations, \(\bar{X}\) and \(\bar{Y}\) are the sample means, and \(n\) is the number of data points. An \(r\)-value close to \(\pm 1\) indicates a strong linear relationship, while \(r \approx 0\) suggests little or no linear correlation.

To assess the statistical significance of \(r\), we test the null hypothesis that \(X\) and \(Y\) are linearly uncorrelated (\(r=0\)). Under this assumption, the following \(t\)-statistic follows a \(t\)-distribution with \(n-2\) degrees of freedom:
\[
t = \frac{r \sqrt{n-2}}{\sqrt{1 - r^2}}.
\]
The p-value is derived from the \(t\)-distribution:
- A small p-value (e.g., \(p < 0.05\)) indicates that the observed correlation is unlikely due to random chance, providing evidence against the null hypothesis.
- A large p-value suggests insufficient evidence to reject the null hypothesis.

When interpreting \(r\) and \(p\):
- A high absolute value of \(r\) (e.g., \(|r| > 0.5\)) implies a stronger linear correlation.
- A small p-value (e.g., \(p < 0.05\)) denotes statistical significance, meaning the correlation is not easily attributed to random variation.
- If \(p\) is not small, we cannot confidently conclude a significant linear relationship.

Taken together, \(r\) and \(p\) provide insight into both the strength and the statistical reliability of the linear relationship between two variables.

\section{Additional Quantitative Study Details}
\label{appendix:B}
\subsection{Effective Average Drop (EAD) Weighting Function}
\label{appendix: EAD Weighting Function}

To define the area-based weighting function \( f(x) \), we consider the function:
\[
f(x) = a \cdot \frac{x}{1 + kx^{2n}},
\]
where \( x \in [0,1] \) is the proportion of the saliency map's area, and \( a, k, n \) are parameters controlling the shape of the function.

We set the optimal area \( E = 0.25 \), and design \( f(x) \) such that:
\[
f(E) = 1, \quad f'(E) = 0.
\]

Solving these conditions, we obtain one of the solutions:
\[
f(x) = 5 \cdot \frac{x}{1 + 256x^5}.
\]

\begin{figure}[H]
    \centering
    \includegraphics[width=0.2\textwidth]{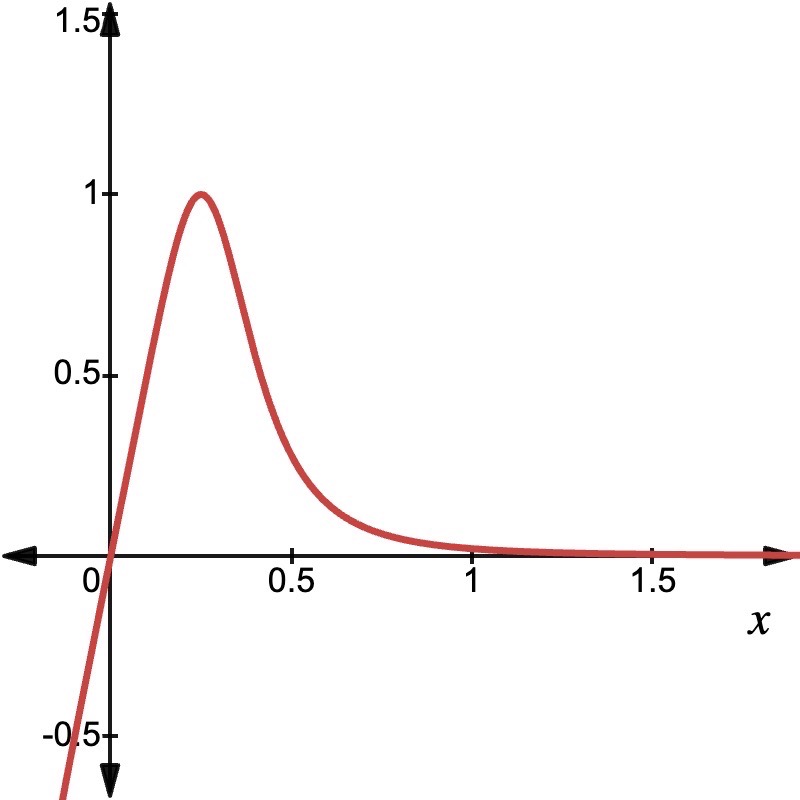} 
    \caption{The weighting function \( f(x) = 5 \cdot \frac{x}{1 + 256x^5} \), which maximizes evaluation when the saliency map covers approximately 25\% of the input image area.}
    \label{fig:area_weight}
\end{figure}

Figure~\ref{fig:area_weight} illustrates the curve of \( f(x) \). The function penalizes saliency maps that are too small or too large, ensuring that the evaluation favors maps covering approximately 25\% of the input image area.

\subsection{Insertion and Deletion Curves}

For each sample, the saliency score \( S_j \) is used to rank pixels by their importance. The model's confidence is measured as pixels are progressively inserted or deleted in increments.

The average value at each step \( i \) is computed as:
\[
\text{Avg Curve}_i = \frac{1}{N} \sum_{j=1}^{N} y_j(f(I_j, S_j^{(i)})),
\]
where:
\begin{itemize}
    \item \( I_j \): Input image for the \( j \)-th sample.
    \item \( y_j \): True label for the \( j \)-th sample.
    \item \( S_j^{(i)} \): The modified image at step \( i \), with pixels inserted or deleted based on the saliency score.
    \item \( f \): The function representing the insertion or deletion operation.
\end{itemize}

By plotting the average confidence over all samples at each step, we obtain the insertion and deletion curves. The Area Under the Curve (AUC) is then computed to quantify the overall performance.

\subsection{Results of Insertion and Deletion Curves}
\label{appendix:Results of Insertion and Deletion Curves}

We employ the \textbf{Insertion} and \textbf{Deletion} curves to evaluate the contribution of saliency scores to model predictions~\cite{petsiuk2018rise}. The insertion curve measures the model's confidence recovery as salient regions are progressively added to a baseline image, while the deletion curve measures the confidence degradation as salient regions are progressively removed.

\begin{figure}[ht]
    \centering
    \begin{minipage}{0.2\textwidth}
        \centering
        \includegraphics[width=\textwidth]{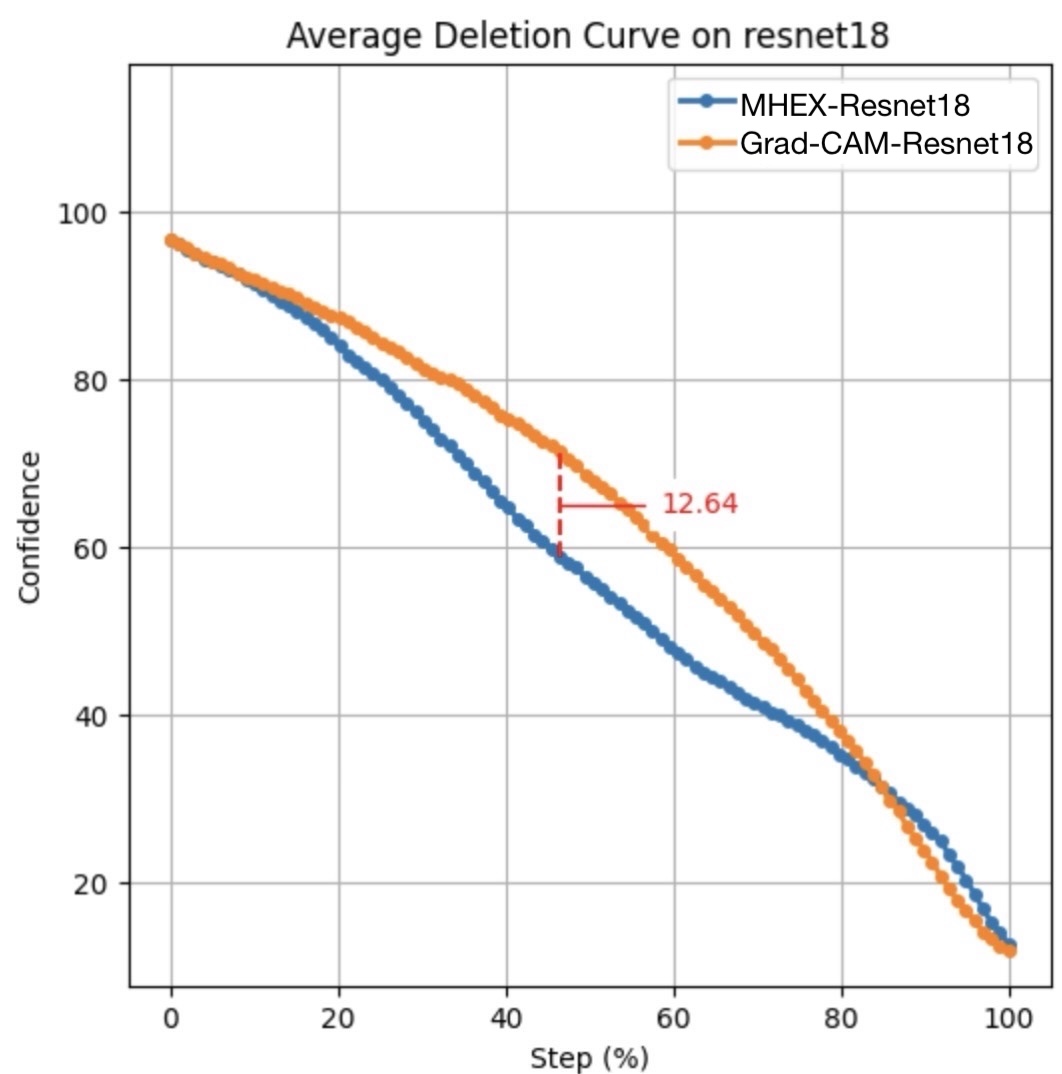}
        \subcaption{Average Deletion Curve on ResNet18}
        \label{fig:avg_del_18}
    \end{minipage}
    \begin{minipage}{0.2\textwidth}
        \centering
        \includegraphics[width=\textwidth]{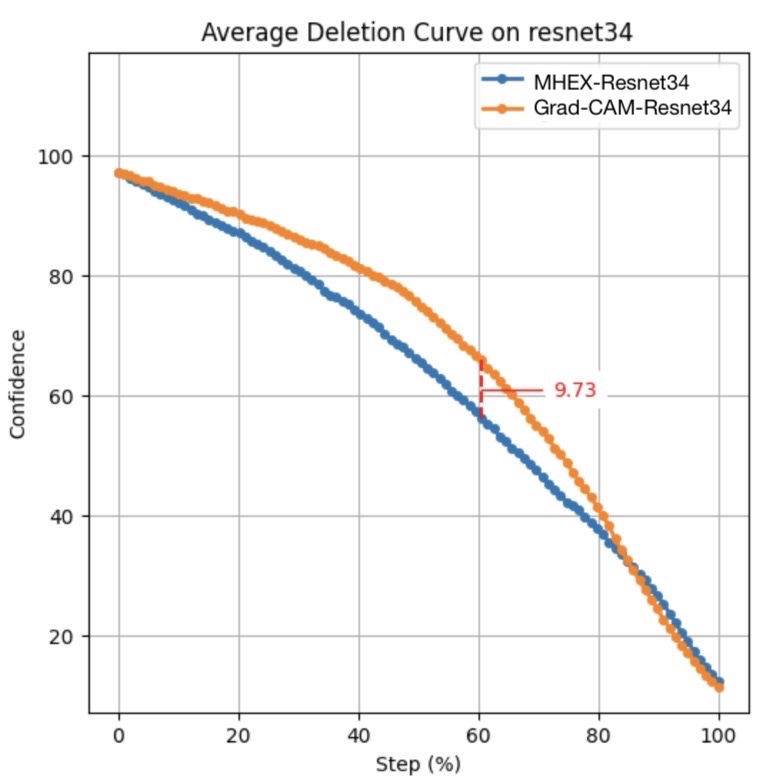}
        \subcaption{Average Deletion Curve on ResNet34}
        \label{fig:avg_del_34}
    \end{minipage}
    \begin{minipage}{0.2\textwidth}
        \centering
        \includegraphics[width=\textwidth]{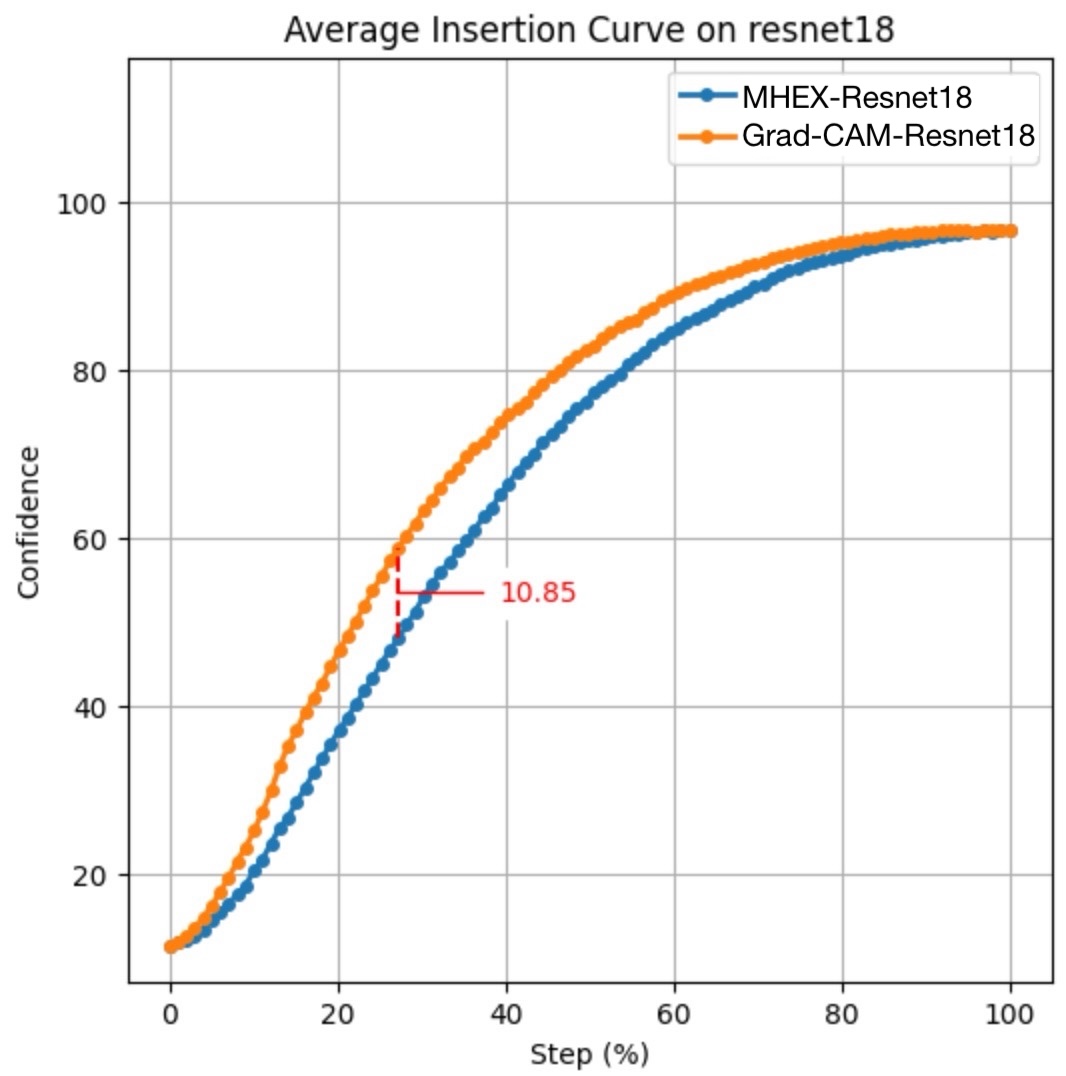}
        \subcaption{Average Insertion Curve on ResNet18}
        \label{fig:avg_ins_18}
    \end{minipage}
    \begin{minipage}{0.2\textwidth}
        \centering
        \includegraphics[width=\textwidth]{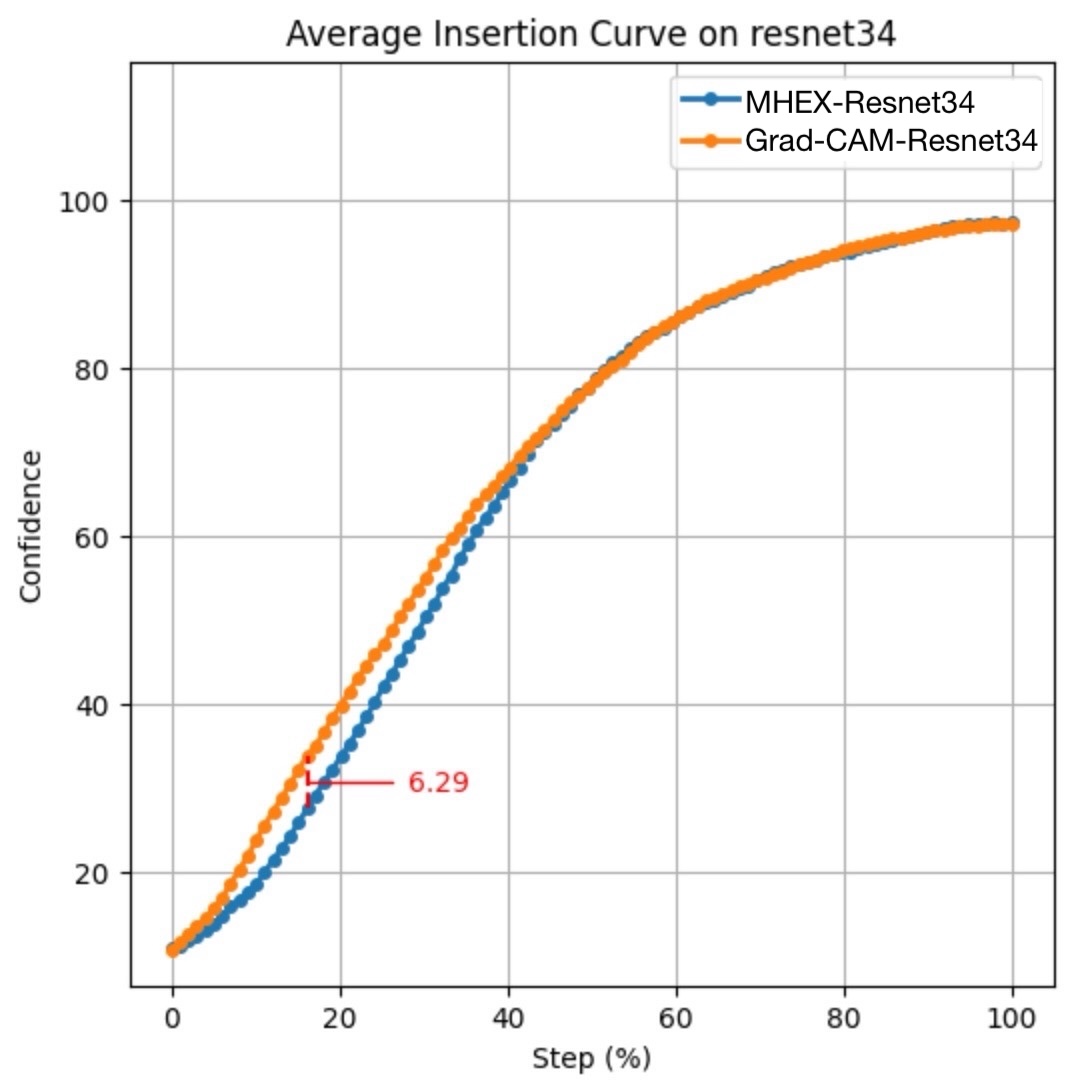}
        \subcaption{Average Insertion Curve on ResNet34}
        \label{fig:avg_ins_34}
    \end{minipage}
    \caption{Comparison of insertion and deletion curves for \textbf{MHEX-Net} and Grad-CAM on ResNet18 and ResNet34. Insertion curves reflect the model’s confidence recovery when saliency regions are added, while deletion curves show confidence drops when regions are removed. \textbf{MHEX-Net}'s dispersed saliency causes slower recovery in insertion curves but sharper drops in deletion curves.}
    \label{fig:insertion-deletion-curves}
\end{figure}

Insertion and deletion curves assess the impact of saliency scores on model confidence. The insertion curve tracks confidence recovery as salient regions are added to a blank image, with AUC reflecting their positive contribution (Figures~\ref{fig:avg_ins_18} and \ref{fig:avg_ins_34}). The deletion curve measures confidence degradation as salient regions are removed, with AUC quantifying their importance (Figures~\ref{fig:avg_del_18} and \ref{fig:avg_del_34}).

Results show that \textbf{MHEX-Net} saliency scores, though slower in confidence recovery during insertion due to broader dispersion, lead to sharper confidence drops during deletion. This highlights \textbf{MHEX-Net}'s superior coverage of critical regions compared to Grad-CAM, which focuses on compact high-confidence areas and may overlook secondary features.

\begin{figure}[H]
\centering
\includegraphics[width=0.45\textwidth]{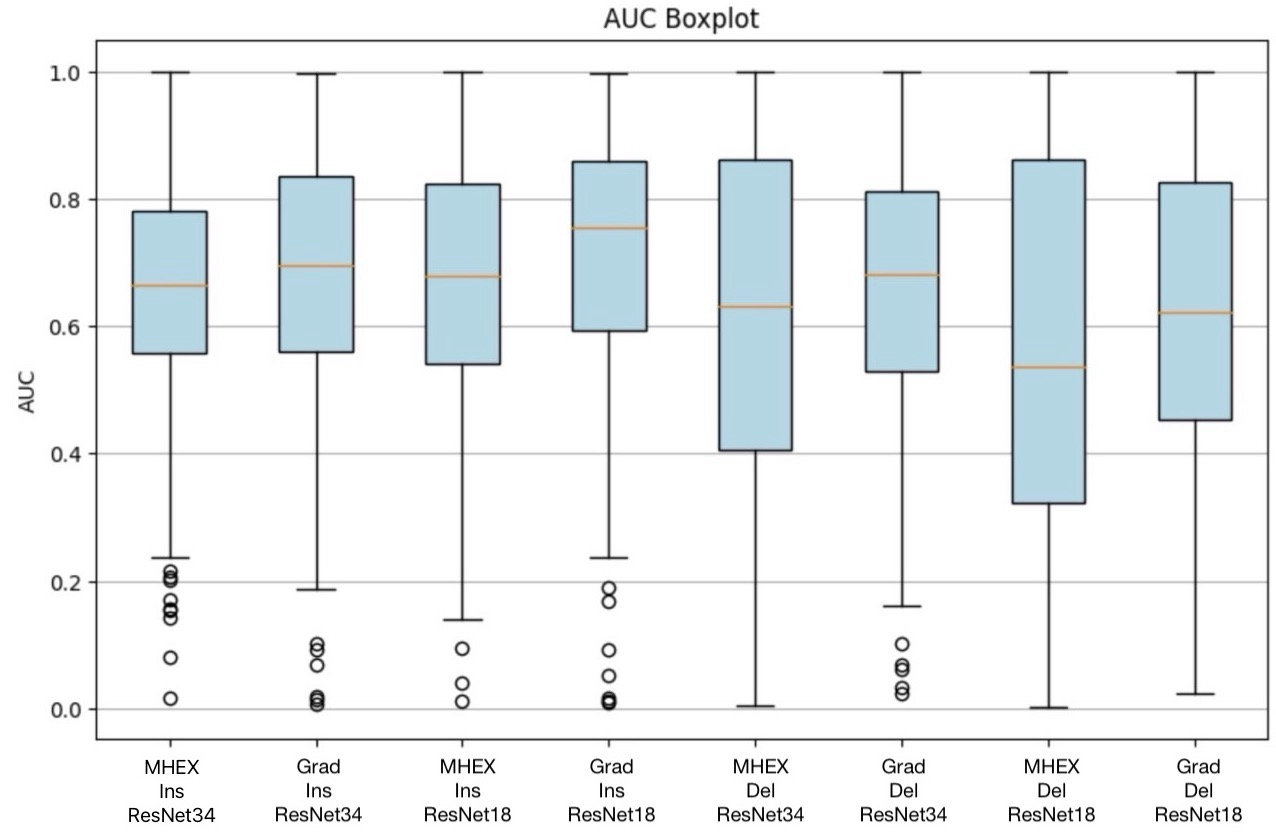}
\caption{AUC Boxplots for Insertion and Deletion on ResNet18 and ResNet34.}
\label{fig:auc-boxplots}
\end{figure}

\begin{figure}[H]
    \centering
    \setlength{\arrayrulewidth}{0pt}
    \begin{tabular}{ccc}
        {\hbox{\includegraphics[width=0.13\textwidth]{images_jpeg/colorectal_epithelium_DEN.jpeg}}} & 
        {\hbox{\includegraphics[width=0.15\textwidth]{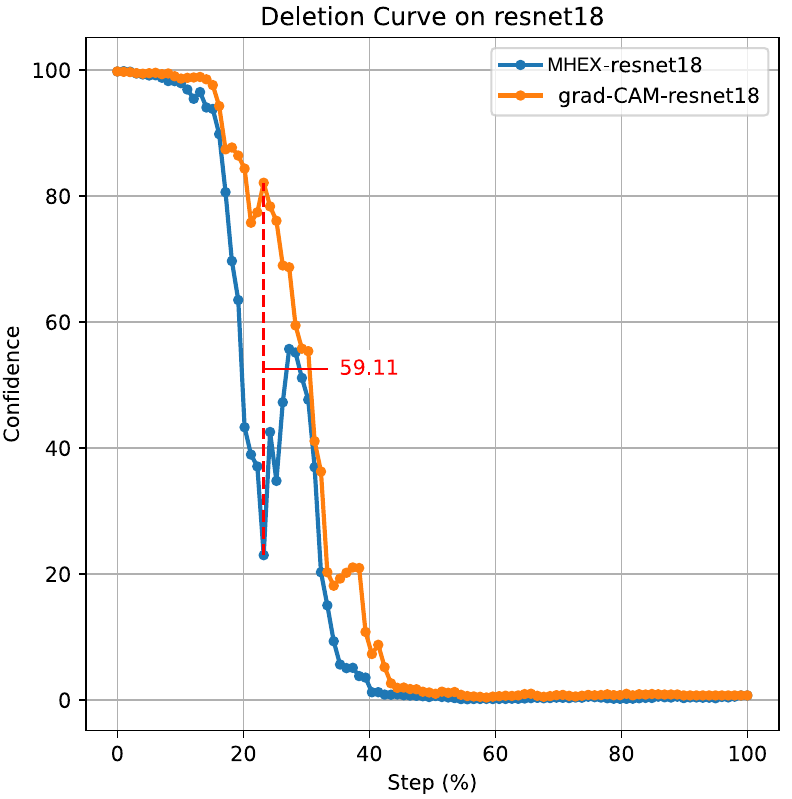}}} &  
        {\hbox{\includegraphics[width=0.15\textwidth]{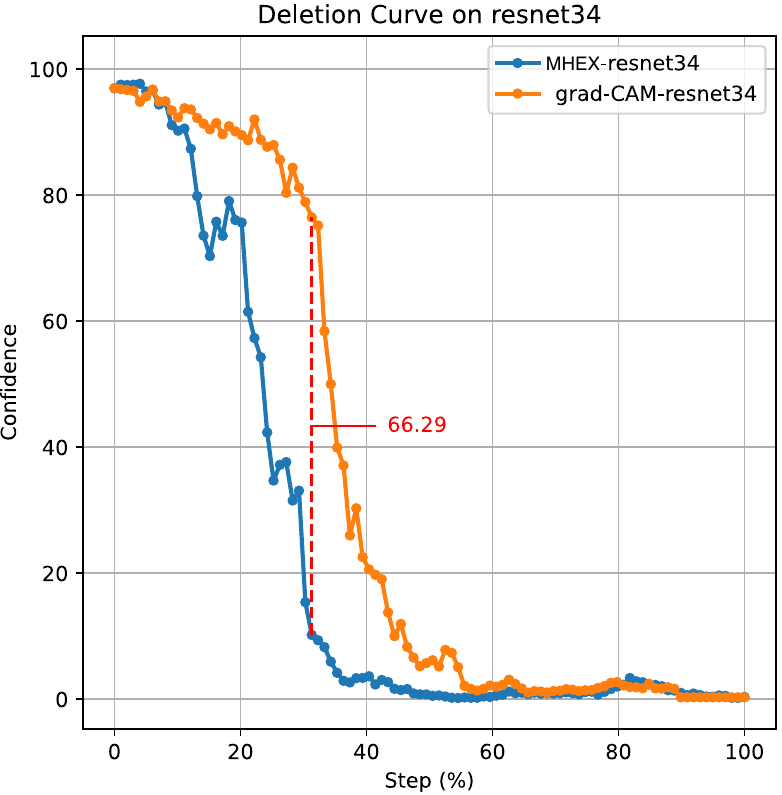}}} \\ 
        {\hbox{\includegraphics[width=0.13\textwidth]{images_jpeg/colorectal_epithelium_gradCAM.jpeg}}} & 
        {\hbox{\includegraphics[width=0.15\textwidth]{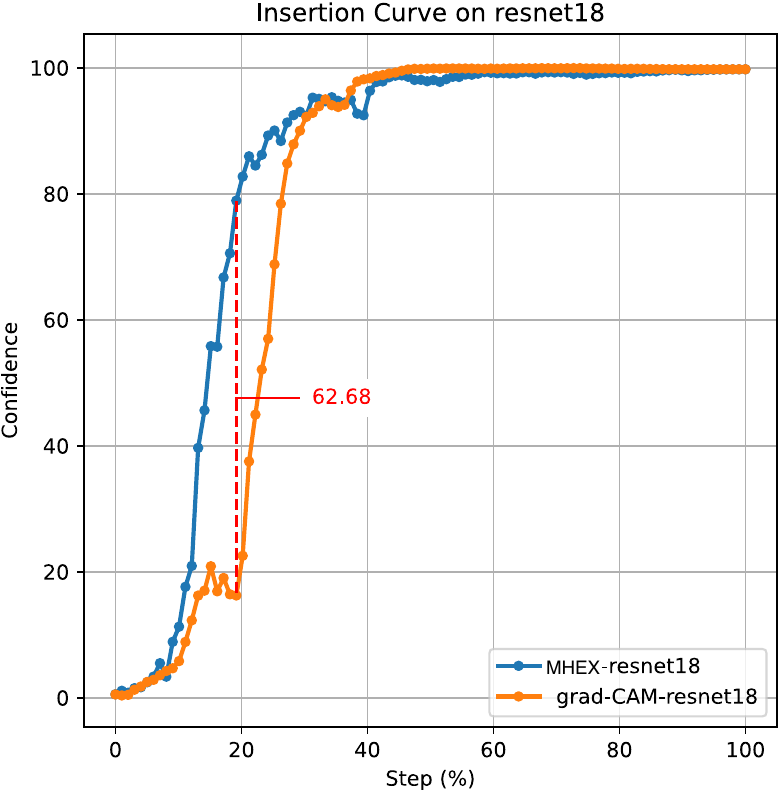}}} &  
        {\hbox{\includegraphics[width=0.15\textwidth]{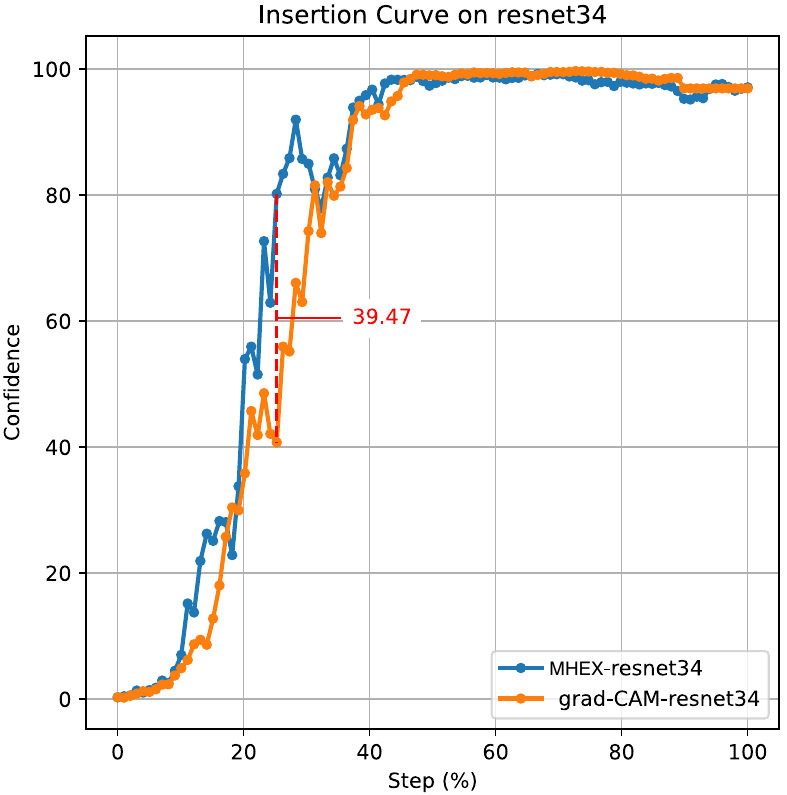}}} \\

        {\hbox{\includegraphics[width=0.13\textwidth]{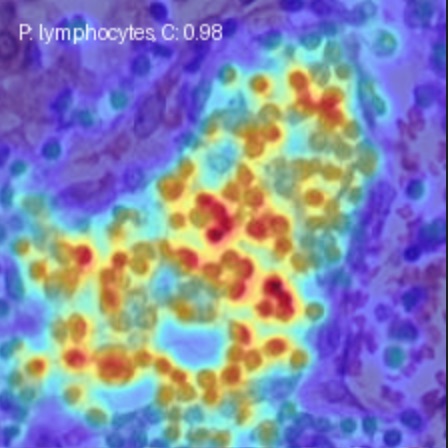}}} & 
        {\hbox{\includegraphics[width=0.15\textwidth]{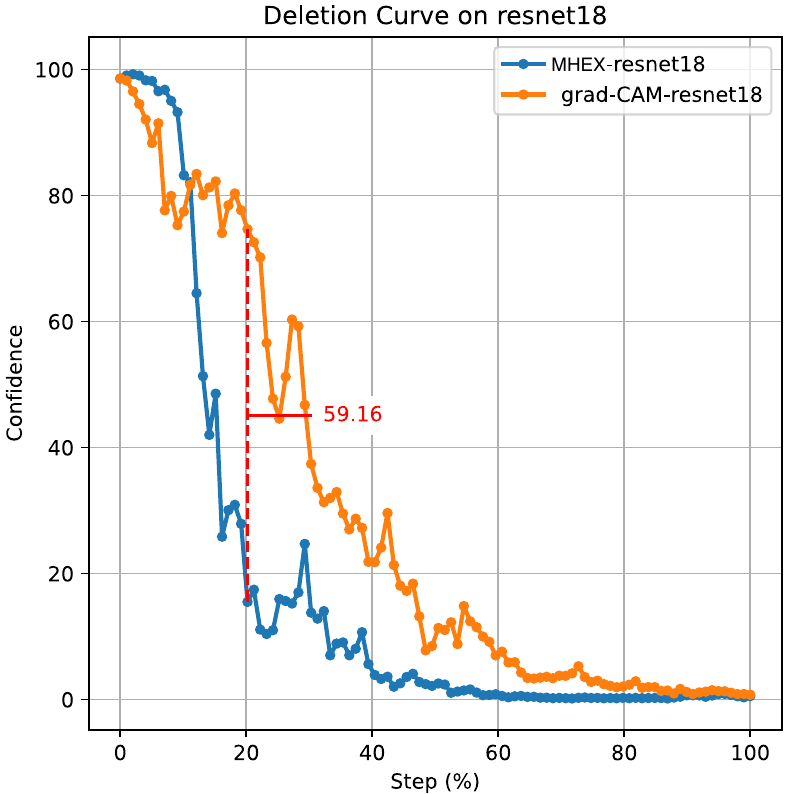}}} &  
        {\hbox{\includegraphics[width=0.15\textwidth]{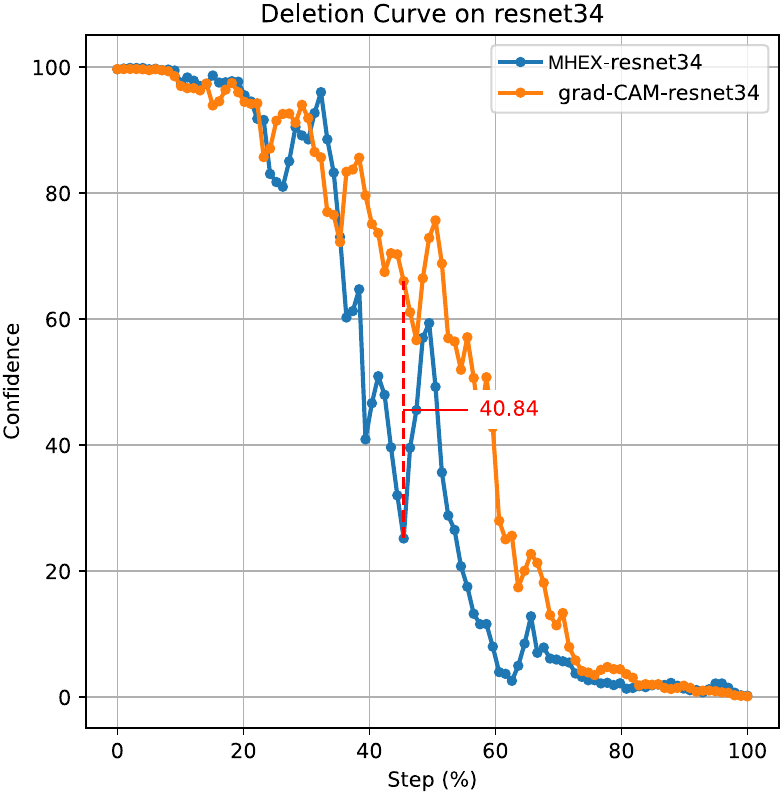}}} \\ 
        {\hbox{\includegraphics[width=0.13\textwidth]{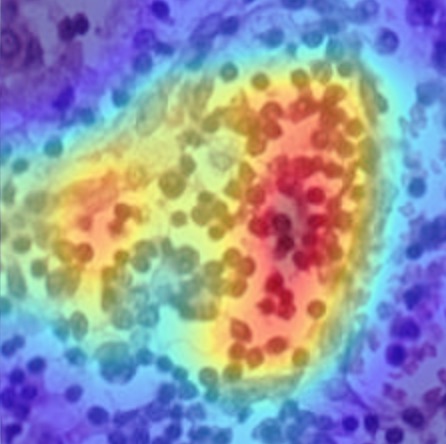}}} & 
        {\hbox{\includegraphics[width=0.15\textwidth]{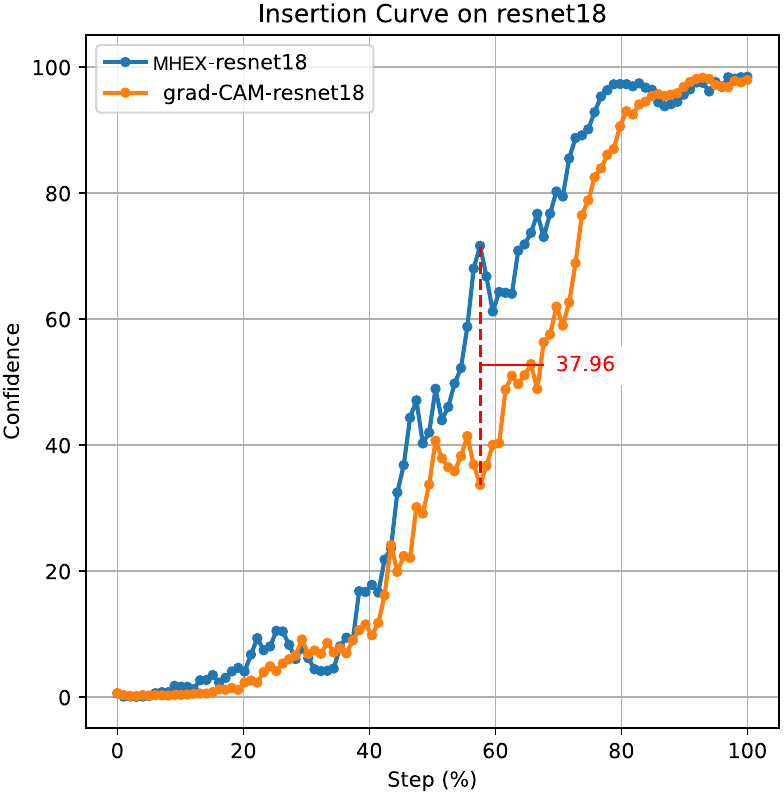}}} &  
        {\hbox{\includegraphics[width=0.15\textwidth]{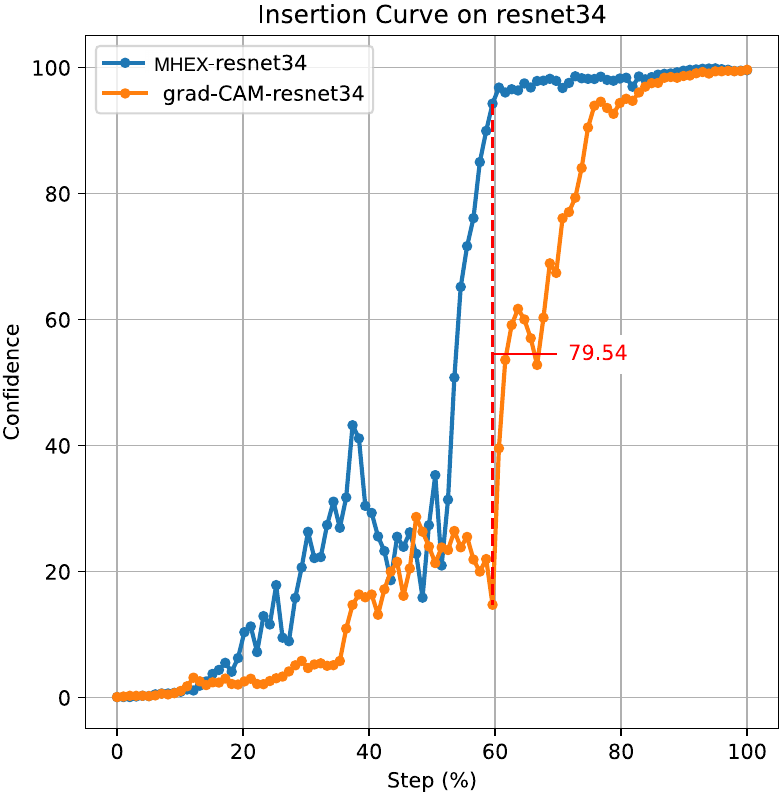}}} \\ 
    \end{tabular}
    \caption{Saliency Score Comparison with Insertion and Deletion Curves.}
    \label{tab:non_negativity}
\end{figure}

\section{More Results}
\label{appendix:C}

To showcase the effectiveness of MHEX-Net across diverse datasets, we provide additional saliency map visualizations. These results, shown in 
Figure~\ref{tab:imagenet1k_den_gradcam_comparison} (ImageNet 1k),
Figure~\ref{tab:tissuemnist} (TissueMNIST),
Figure~\ref{tab:pathmnist} (PathMNIST) and , 
Figure~\ref{tab:bloodmnist} (BloodMNIST), demonstrate that MHEX-enhanced ResNet models exhibit remarkable sensitivity to medical details, achieving near-semantic segmentation accuracy for most samples. This superior interpretability is particularly evident in challenging datasets, as highlighted by these examples, underscoring the robustness and precision of MHEX in capturing critical features.
Additionally, more saliency scores for AG News can be found in Table~\ref{tab:salience_transformer_extra}.
\begin{table*}[!h]
    \centering
    \setlength{\arrayrulewidth}{0pt}
    \renewcommand{\arraystretch}{1.4} 
   \includegraphics[width=\textwidth]{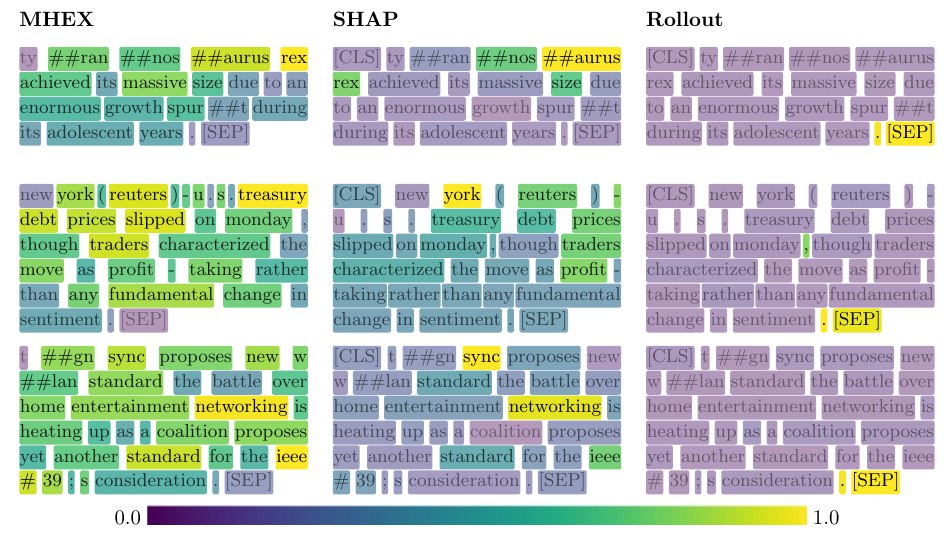}
    \caption{Saliency Scores Comparison for Selected AG News Samples.}
    \label{tab:salience_transformer_extra}
\end{table*}

\begin{figure}[h!]
    \centering
    \setlength{\arrayrulewidth}{0pt}
    \setlength{\tabcolsep}{1pt} 
    \renewcommand{\arraystretch}{1} 
    \begin{tabular}{|>{\centering\arraybackslash}m{0.02\textwidth}|c|c|c|}
        \hline
         & \textbf{Original Image} & \textbf{MHEX} & \textbf{Grad-CAM} \\
        \hline



        \rotatebox[origin=c]{90}{\textbf{Parachute}} &
        \raisebox{-0.5\height}{\includegraphics[width=0.15\textwidth]{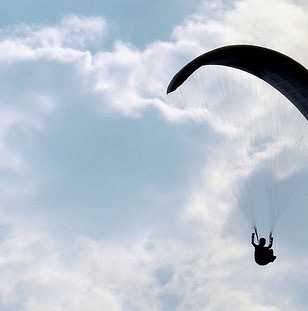}} &
        \raisebox{-0.5\height}{\includegraphics[width=0.15\textwidth]{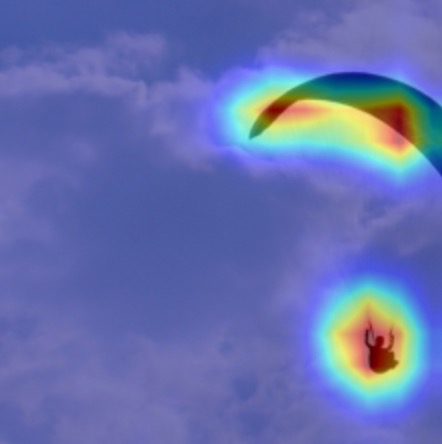}} &
        \raisebox{-0.5\height}{\includegraphics[width=0.15\textwidth]{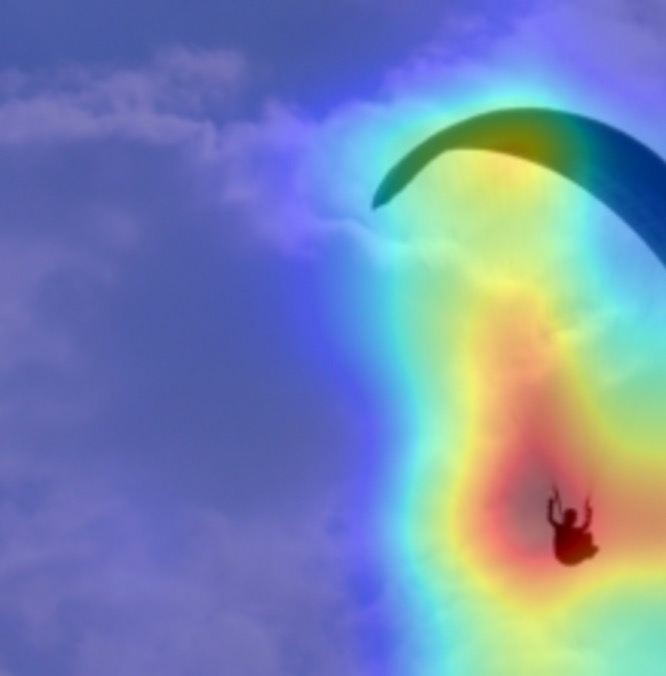}} \\
        \hline


        \rotatebox[origin=c]{90}{\textbf{Vine-snake}} &
        \raisebox{-0.5\height}{\includegraphics[width=0.15\textwidth]{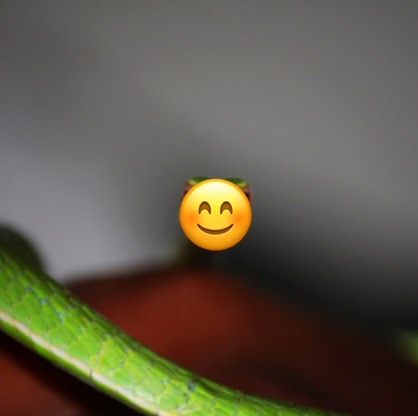}} &
        \raisebox{-0.5\height}{\includegraphics[width=0.15\textwidth]{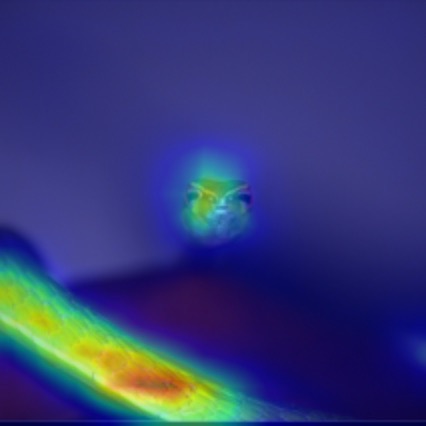}} &
        \raisebox{-0.5\height}{\includegraphics[width=0.15\textwidth]{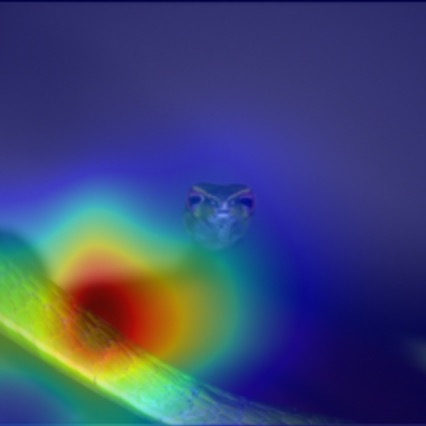}} \\
        \hline

 \rotatebox[origin=c]{90}{\textbf{Book Jacket}} &
        \raisebox{-0.5\height}{\includegraphics[width=0.15\textwidth]{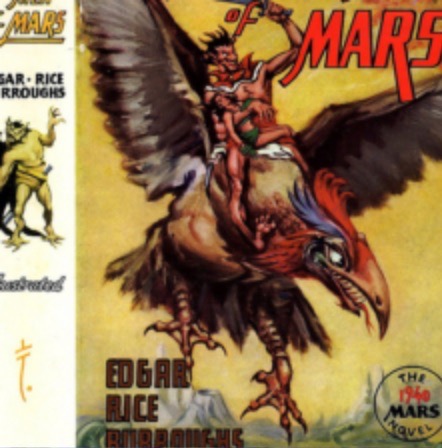}} &
        \raisebox{-0.5\height}{\includegraphics[width=0.15\textwidth]{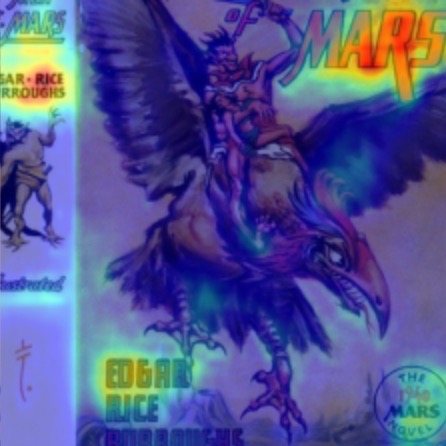}} &
        \raisebox{-0.5\height}{\includegraphics[width=0.15\textwidth]{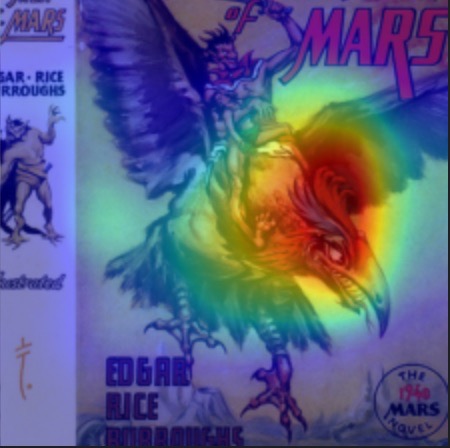}} \\
        \hline

        \rotatebox[origin=c]{90}{\textbf{Leopard}} &
        \raisebox{-0.5\height}{\includegraphics[width=0.15\textwidth]{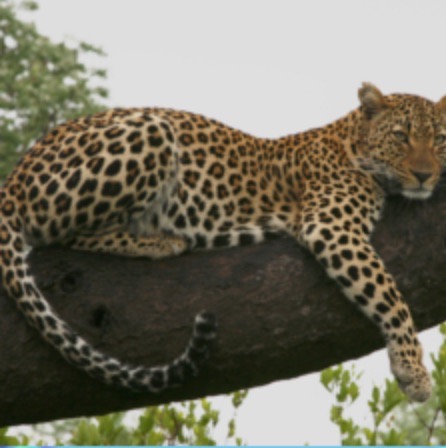}} &
        \raisebox{-0.5\height}{\includegraphics[width=0.15\textwidth]{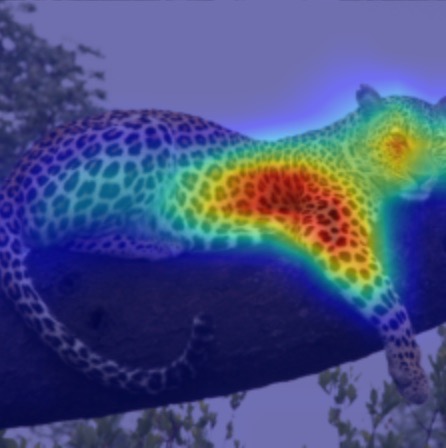}} &
        \raisebox{-0.5\height}{\includegraphics[width=0.15\textwidth]{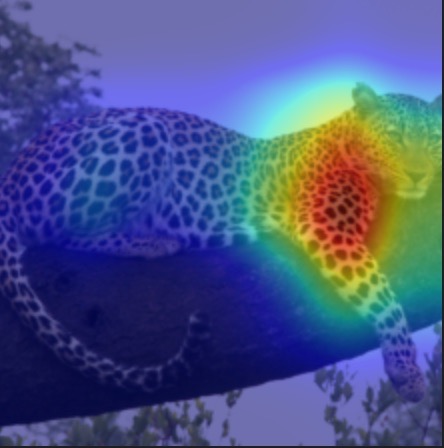}} \\
        \hline

        \rotatebox[origin=c]{90}{\textbf{Pirate}} &
        \raisebox{-0.5\height}{\includegraphics[width=0.15\textwidth]{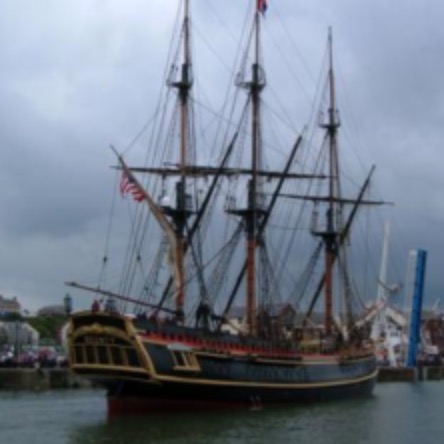}} &
        \raisebox{-0.5\height}{\includegraphics[width=0.15\textwidth]{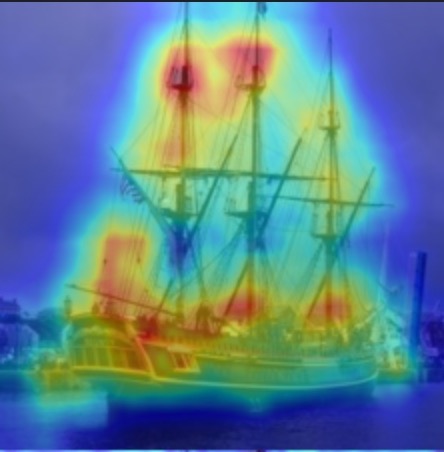}} &
        \raisebox{-0.5\height}{\includegraphics[width=0.15\textwidth]{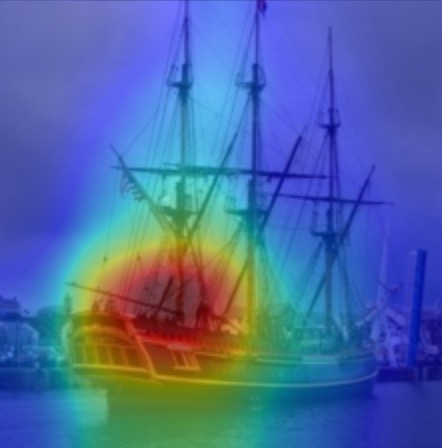}} \\
        \hline

        
    \end{tabular}
    \caption{Comparison of original images, MHEX, and Grad-CAM saliency maps on selected ImageNet1k classes.}
    \label{tab:imagenet1k_den_gradcam_comparison}
\end{figure}

\begin{figure}[ht]
    \centering
    \setlength{\arrayrulewidth}{0pt} 
    \setlength{\tabcolsep}{2pt} 
    \renewcommand{\arraystretch}{1.5} 
    \begin{tabular}{|>{\centering\arraybackslash}m{0.07\textwidth}|>{\centering\arraybackslash}m{0.13\textwidth}|>{\centering\arraybackslash}m{0.13\textwidth}|>{\centering\arraybackslash}m{0.13\textwidth}|}
        \hline
        & \textbf{TissueMNIST} & \textbf{MHEX} & \textbf{Grad-CAM} \\ \hline
        \rotatebox[origin=c]{90}{\makecell{\textbf{Collecting Duct}\\\textbf{Connecting}\\\textbf{Tubule}}} &
        \includegraphics[width=0.13\textwidth]{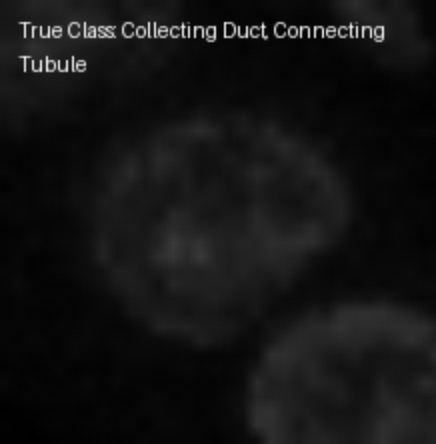} &
        \includegraphics[width=0.13\textwidth]{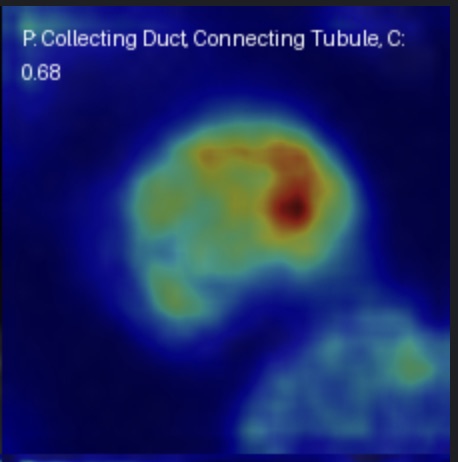} &
        \includegraphics[width=0.13\textwidth]{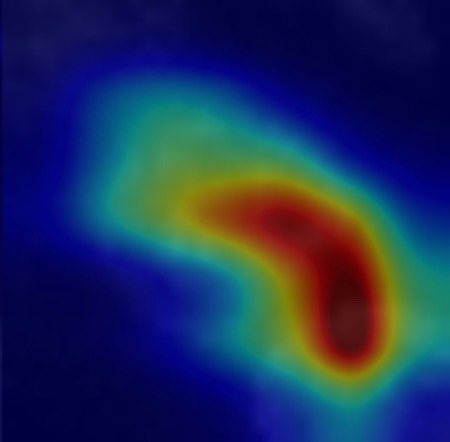} \\ \hline
        \rotatebox[origin=c]{90}{\textbf{Leukocytes}} &
        \includegraphics[width=0.13\textwidth]{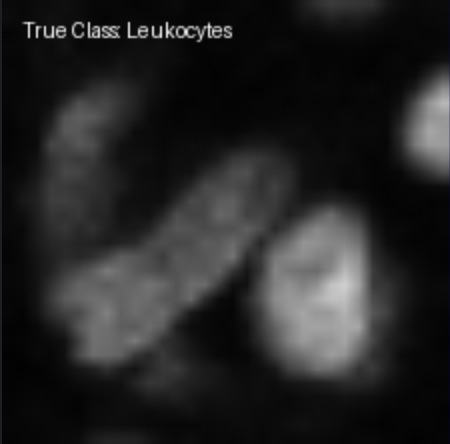} &
        \includegraphics[width=0.13\textwidth]{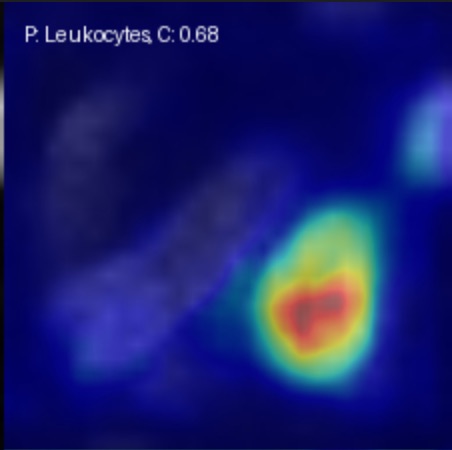} &
        \includegraphics[width=0.13\textwidth]{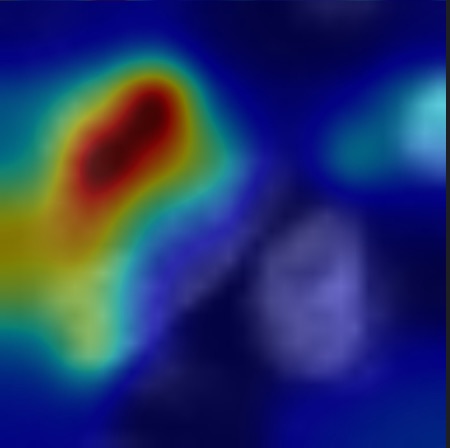} \\ \hline
        \rotatebox[origin=c]{90}{\makecell{\textbf{Glomerular}\\\textbf{Endothelial}\\\textbf{Cells}}} &
        \includegraphics[width=0.13\textwidth]{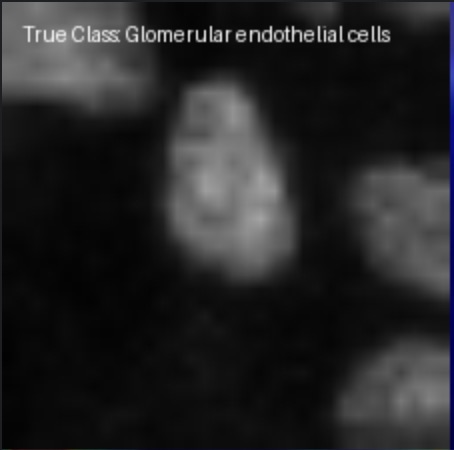} &
        \includegraphics[width=0.13\textwidth]{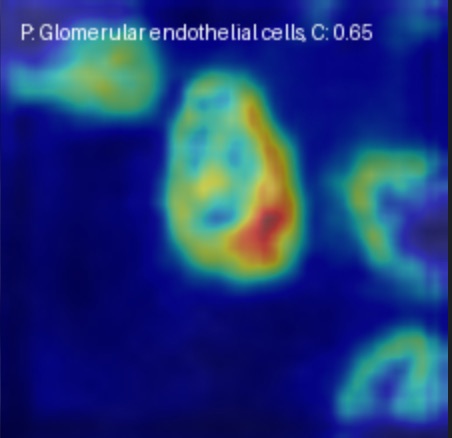} &
        \includegraphics[width=0.13\textwidth]{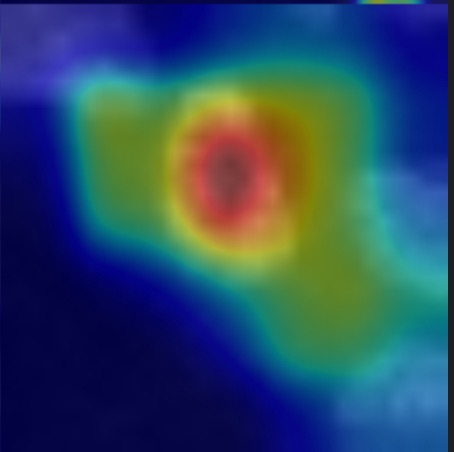} \\ \hline
        \rotatebox[origin=c]{90}{\makecell{\textbf{Distal}\\\textbf{Convoluted}\\\textbf{Tubule}}} &
        \includegraphics[width=0.13\textwidth]{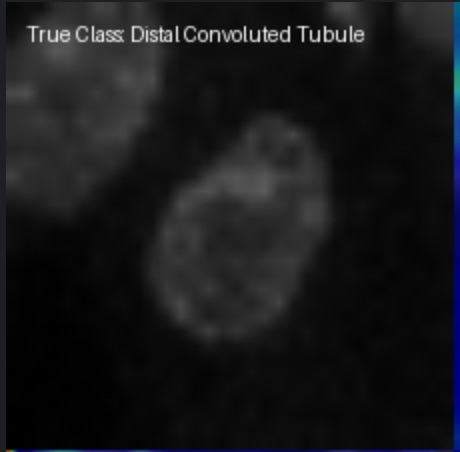} &
        \includegraphics[width=0.13\textwidth]{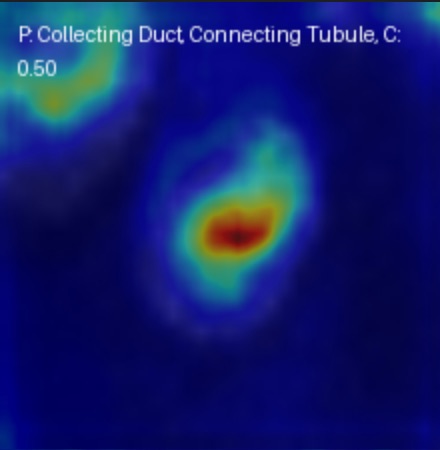} &
        \includegraphics[width=0.13\textwidth]{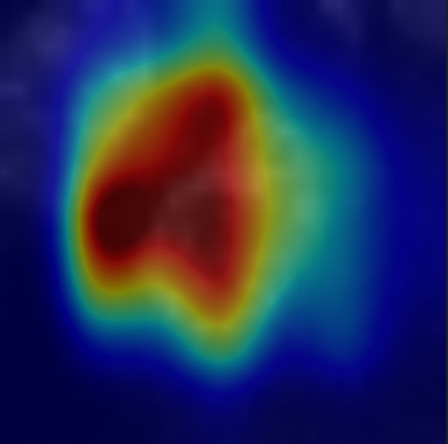} \\ \hline
        \rotatebox[origin=c]{90}{\makecell{\textbf{Thick}\\\textbf{Ascending}\\\textbf{Limb}}} &
        \includegraphics[width=0.13\textwidth]{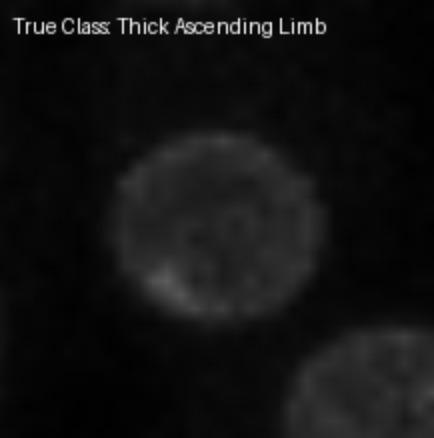} &
        \includegraphics[width=0.13\textwidth]{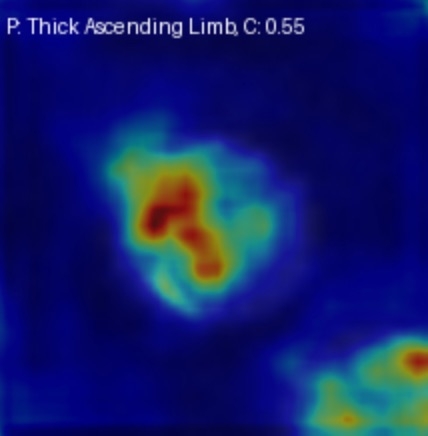} &
        \includegraphics[width=0.13\textwidth]{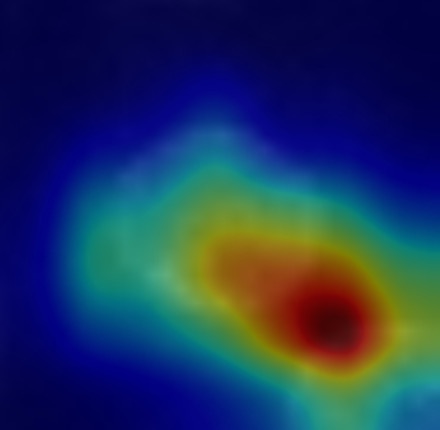} \\ \hline
    \end{tabular}
    \caption{Comparison of saliency maps for TissueMNIST. MHEX showcases a high degree of sensitivity to tissue structures.}
    \label{tab:tissuemnist}
\end{figure}

\begin{figure*}[ht]
    \centering
    \setlength{\arrayrulewidth}{0pt} 
    \setlength{\tabcolsep}{2pt} 
    \renewcommand{\arraystretch}{1.5} 
    \begin{tabular}{|>{\centering\arraybackslash}m{0.06\textwidth}|>{\centering\arraybackslash}m{0.18\textwidth}|>{\centering\arraybackslash}m{0.18\textwidth}|>{\centering\arraybackslash}m{0.18\textwidth}|>{\centering\arraybackslash}m{0.18\textwidth}|>{\centering\arraybackslash}m{0.18\textwidth}|}
    \toprule 
        \hline
        & \textbf{PathMNIST} & \textbf{MHEX} & \textbf{ResNet} & \makecell{\textbf{SHAP} \\ \textbf{50k * 256 evals}} & \textbf{Layer CAM} \\ \hline
        \midrule
        \rotatebox[origin=c]{90}{\makecell{\textbf{Cancer-}\\\textbf{Associated}\\\textbf{Stroma}}} &
        \includegraphics[width=0.17\textwidth]{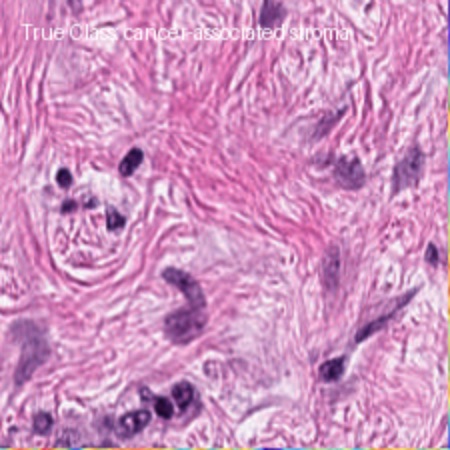} &
        \includegraphics[width=0.17\textwidth]{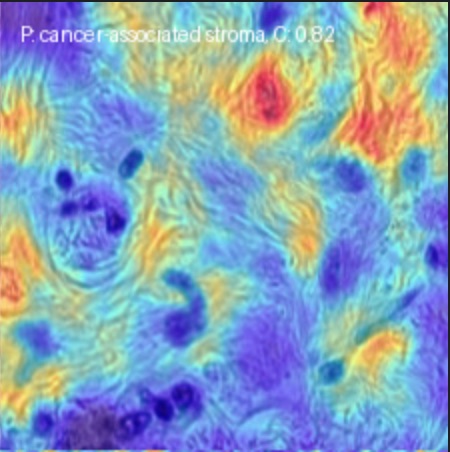} &
        \includegraphics[width=0.17\textwidth]{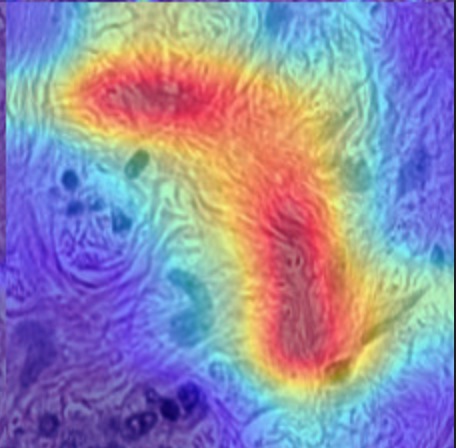} &
        \includegraphics[width=0.17\textwidth]{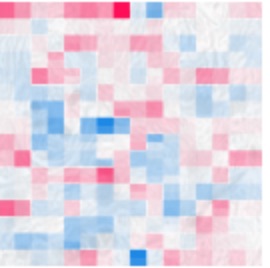} &
        \includegraphics[width=0.17\textwidth]{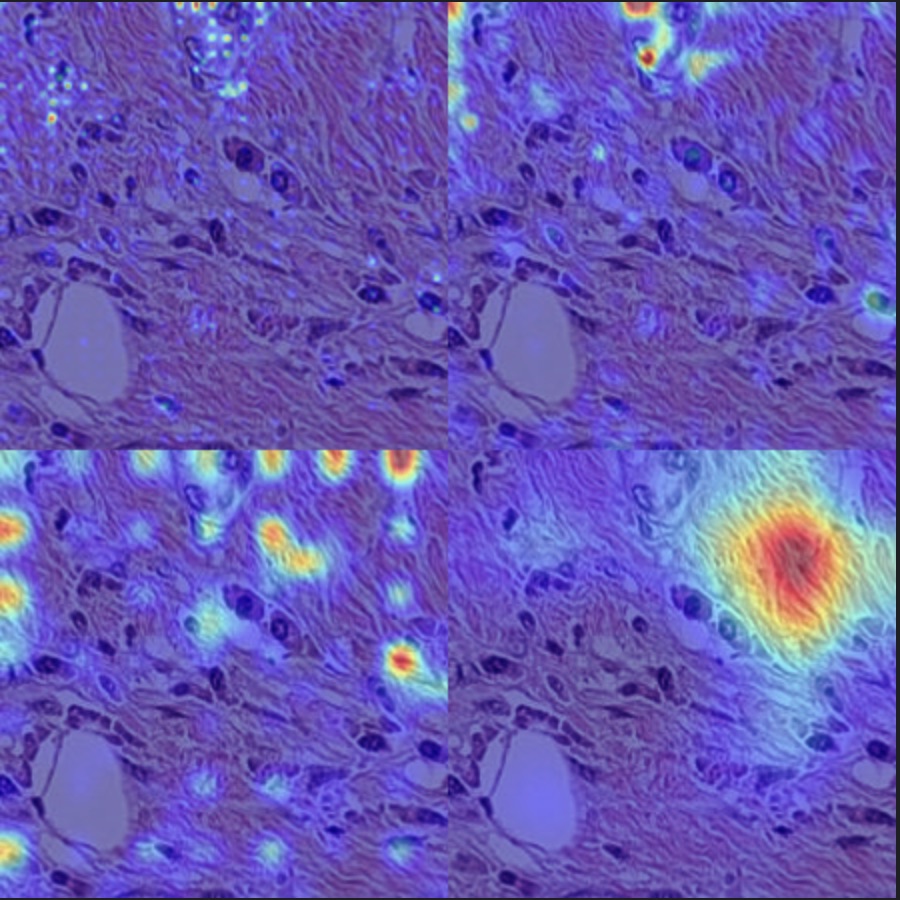} \\ \hline
        \rotatebox[origin=c]{90}{\textbf{Lymphocytes}} &
        \includegraphics[width=0.17\textwidth]{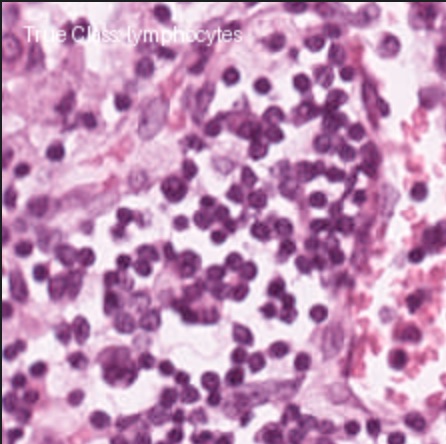} &
        \includegraphics[width=0.17\textwidth]{images_jpeg/appendix/PathMnist_sample2_DEN.jpeg} &
        \includegraphics[width=0.17\textwidth]{images_jpeg/appendix/PathMnist_sample2_ResNet.jpeg} &
        \includegraphics[width=0.17\textwidth]{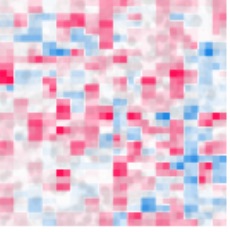} &
        \includegraphics[width=0.17\textwidth]{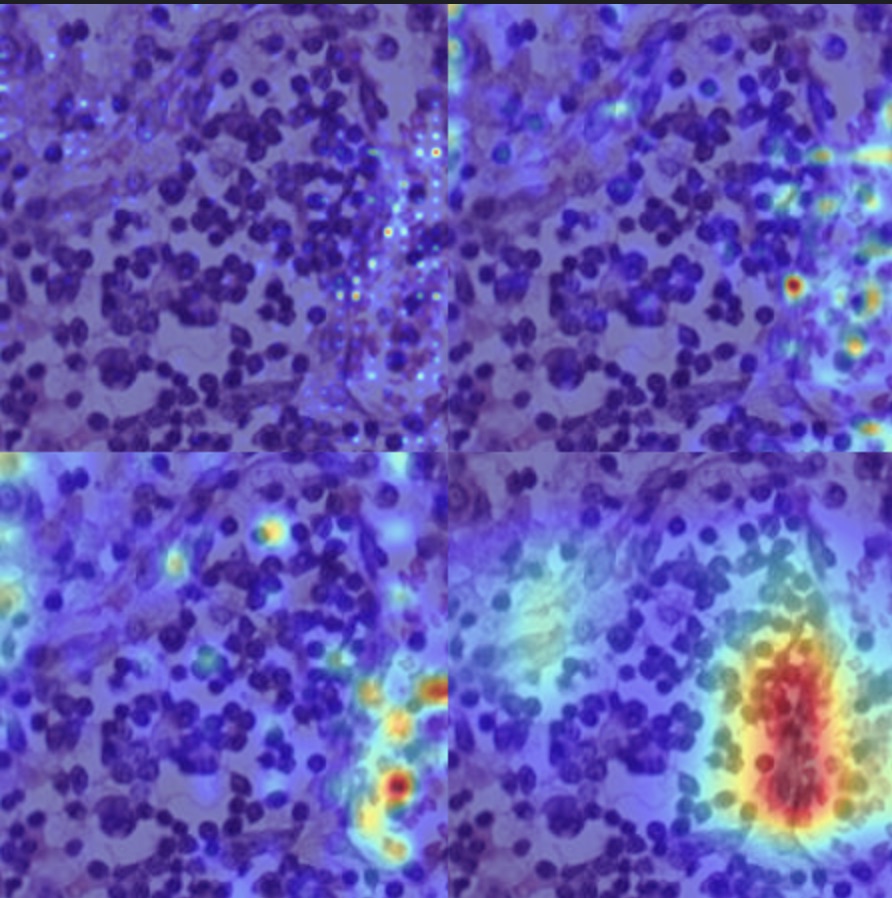} \\ \hline
        \rotatebox[origin=c]{90}{\makecell{\textbf{Colorectal}\\\textbf{Adenocarcinoma}\\\textbf{Epithelium}}} &
        \includegraphics[width=0.17\textwidth]{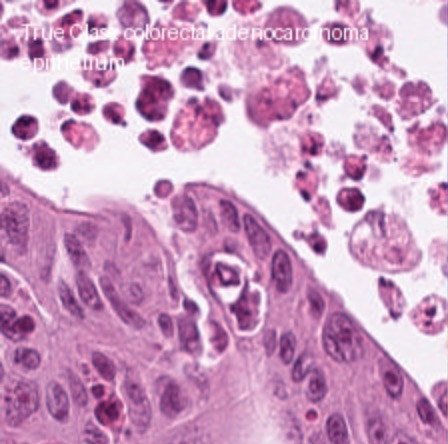} &
        \includegraphics[width=0.17\textwidth]{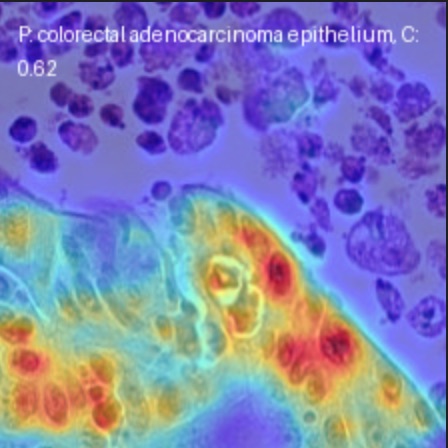} &
        \includegraphics[width=0.17\textwidth]{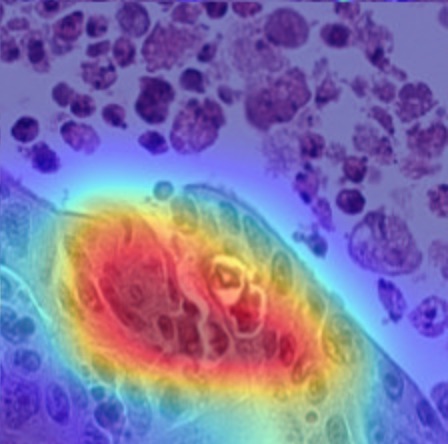} &
        \includegraphics[width=0.17\textwidth]{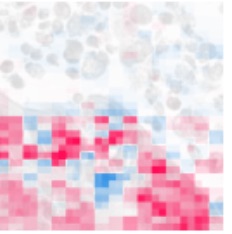} &
        \includegraphics[width=0.17\textwidth]{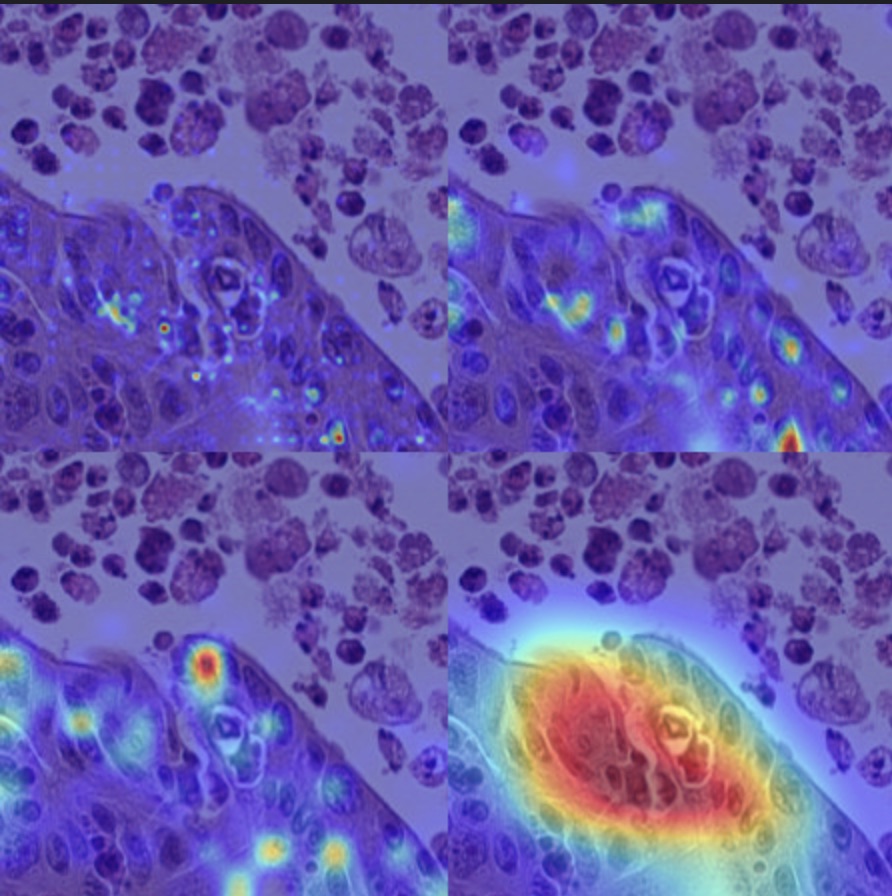} \\ \hline
        \rotatebox[origin=c]{90}{\textbf{Mucus}} &
        \includegraphics[width=0.17\textwidth]{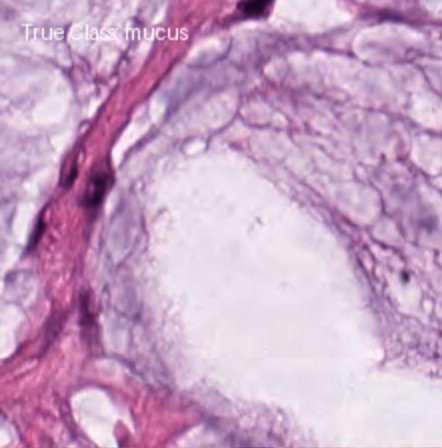} &
        \includegraphics[width=0.17\textwidth]{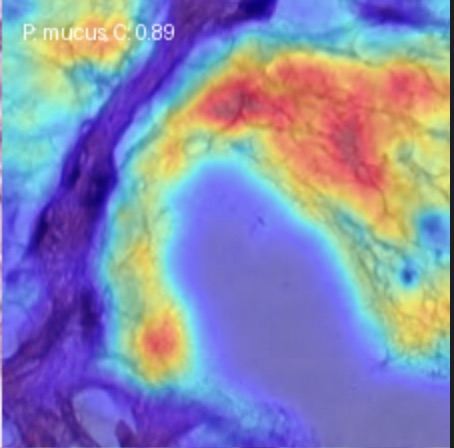} &
        \includegraphics[width=0.17\textwidth]{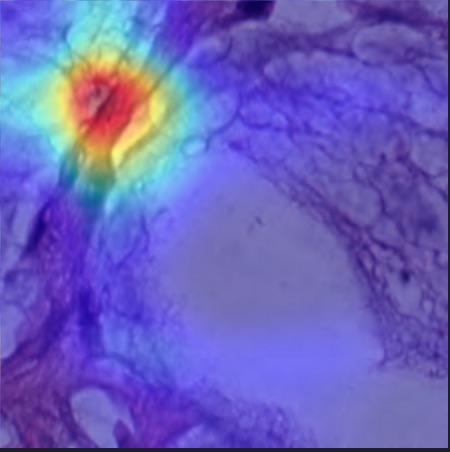} &
        \includegraphics[width=0.17\textwidth]{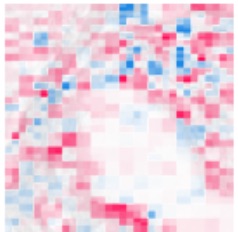} &
        \includegraphics[width=0.17\textwidth]{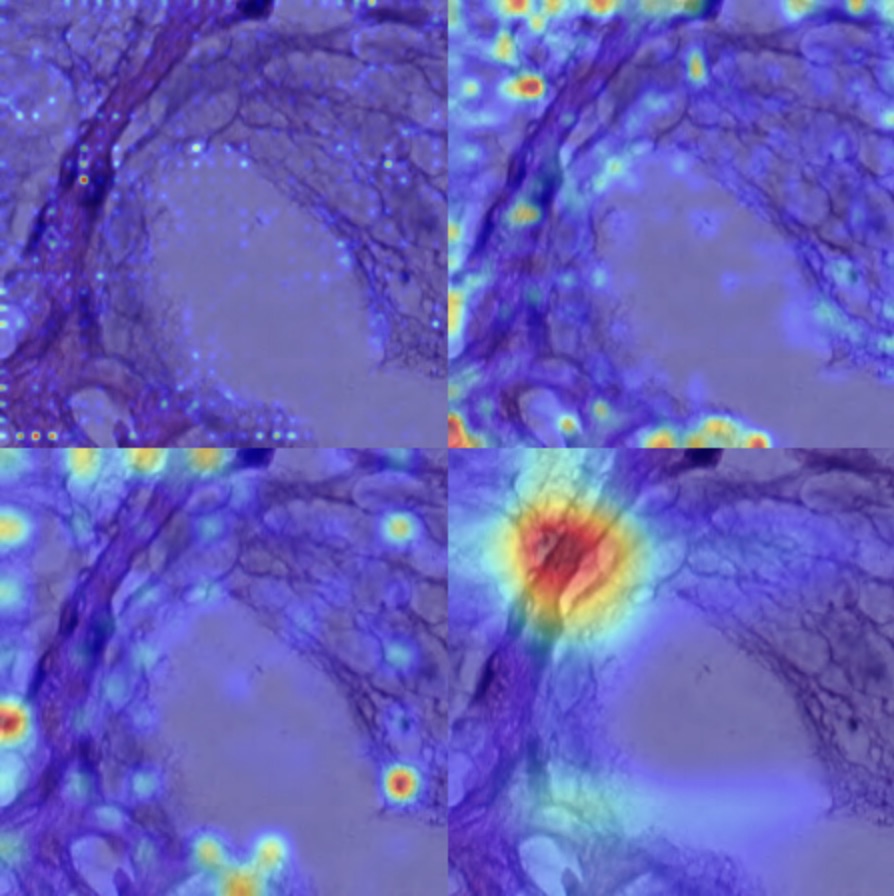} \\ \hline
        \rotatebox[origin=c]{90}{\makecell{\textbf{Smooth}\\\textbf{Muscle}}} &
        \includegraphics[width=0.17\textwidth]{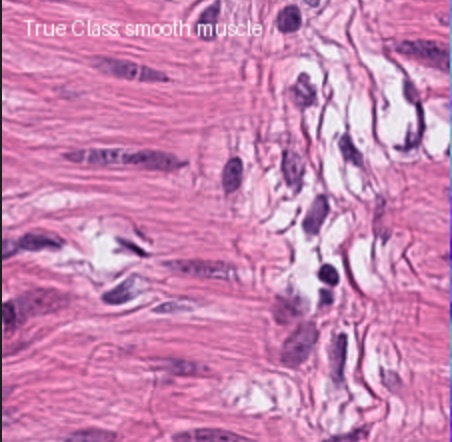} &
        \includegraphics[width=0.17\textwidth]{images_jpeg/appendix/PathMnist_sample5_DEN.jpeg} &
        \includegraphics[width=0.17\textwidth]{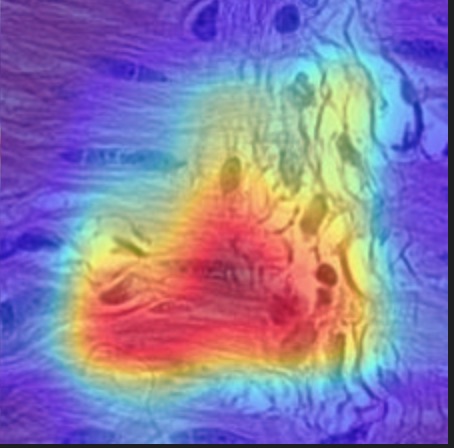} &
        \includegraphics[width=0.17\textwidth]{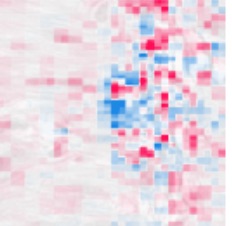} &
        \includegraphics[width=0.17\textwidth]{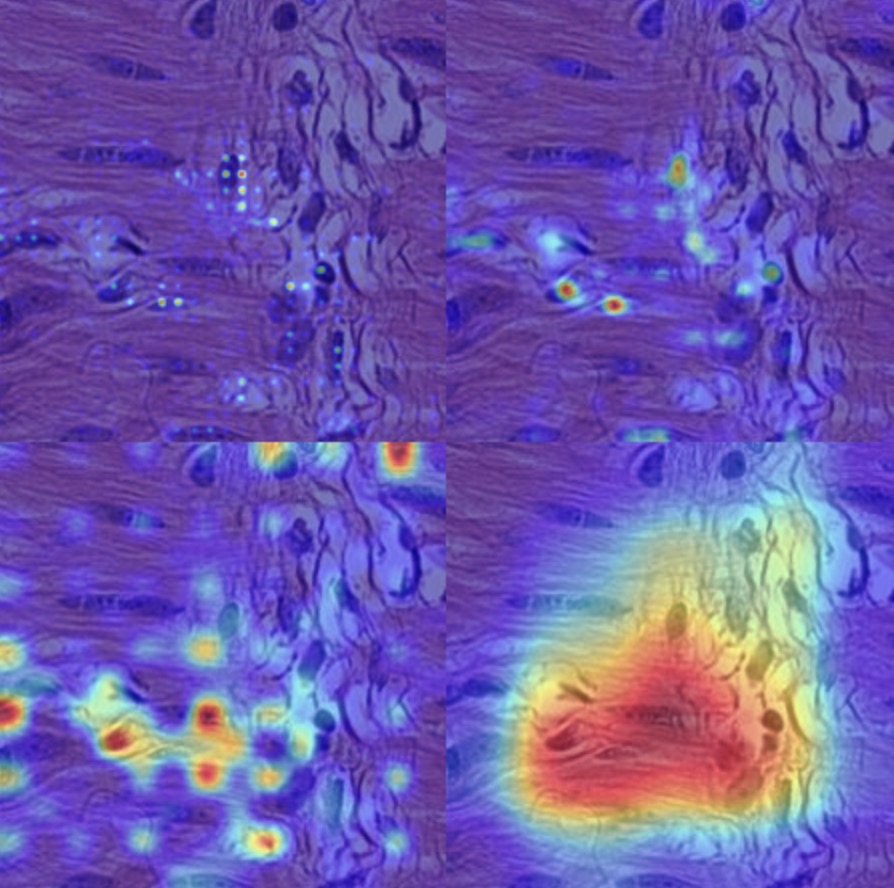} \\ \hline
    \end{tabular}
    \caption{Comparison of saliency maps for PathMNIST. MHEX consistently highlights critical regions with near-segmentation accuracy.}
    \label{tab:pathmnist}
\end{figure*}

\begin{figure*}[ht]
    \centering
    \setlength{\arrayrulewidth}{0pt}
    \setlength{\tabcolsep}{0.1pt} 
    \renewcommand{\arraystretch}{0.1} 
    \begin{tabular}{|>{\centering\arraybackslash}m{0.04\textwidth}|>{\centering\arraybackslash}m{0.18\textwidth}|>{\centering\arraybackslash}m{0.18\textwidth}|>{\centering\arraybackslash}m{0.18\textwidth}|>{\centering\arraybackslash}m{0.18\textwidth}|>{\centering\arraybackslash}m{0.18\textwidth}|}
        \hline
        \toprule
        & \textbf{BloodMNIST} & \textbf{MHEX} & \textbf{Grad-CAM} & \makecell{\textbf{SHAP}\\\textbf{50k * 256 evals}} & \textbf{Layer CAM} \\ \hline
        \midrule
        \rotatebox[origin=c]{90}{\makecell{\textbf{Eosinophil}}} &
        \includegraphics[width=0.17\textwidth]{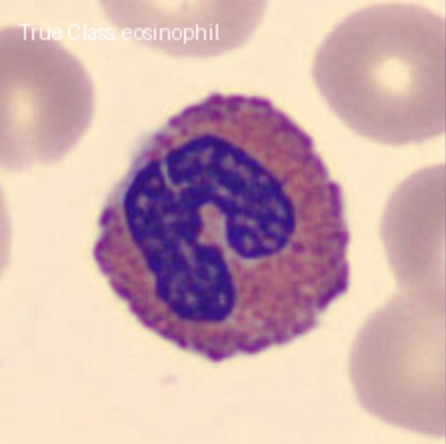} &
        \includegraphics[width=0.17\textwidth]{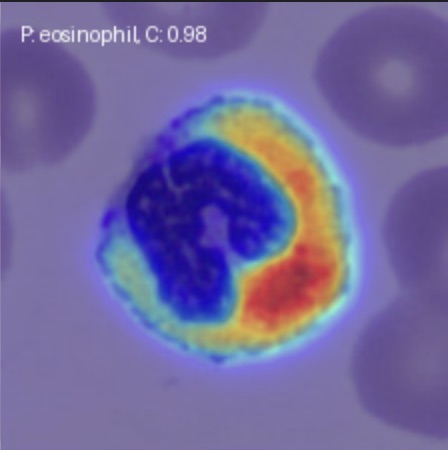} &
        \includegraphics[width=0.17\textwidth]{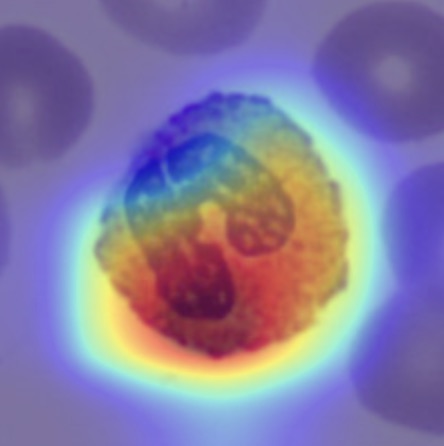} &
        \includegraphics[width=0.17\textwidth]{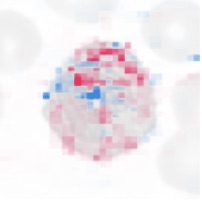} &
        \includegraphics[width=0.17\textwidth]{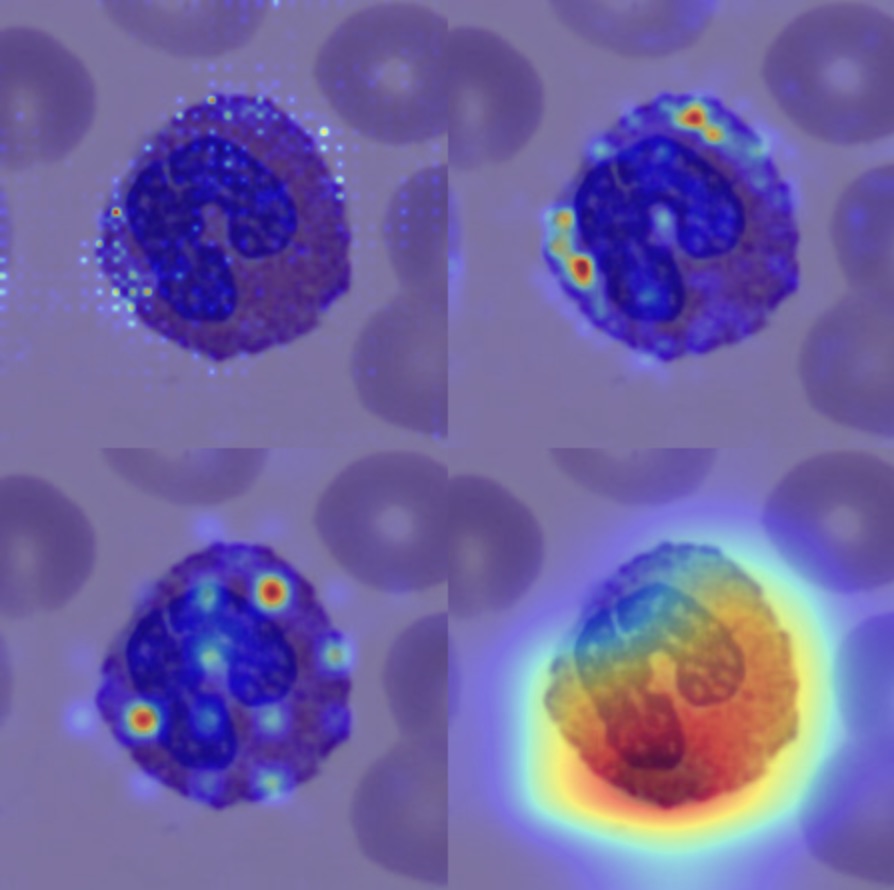} \\ \hline
        \rotatebox[origin=c]{90}{\makecell{\textbf{Basophil}}} &
        \includegraphics[width=0.17\textwidth]{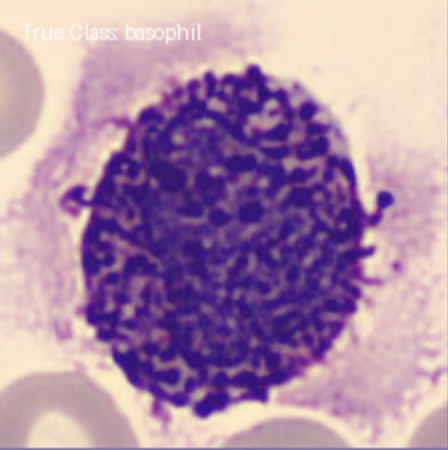} &
        \includegraphics[width=0.17\textwidth]{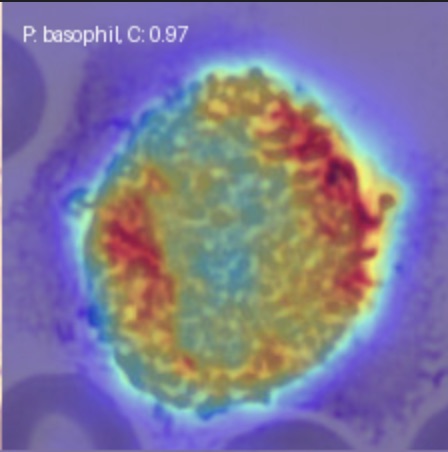} &
        \includegraphics[width=0.17\textwidth]{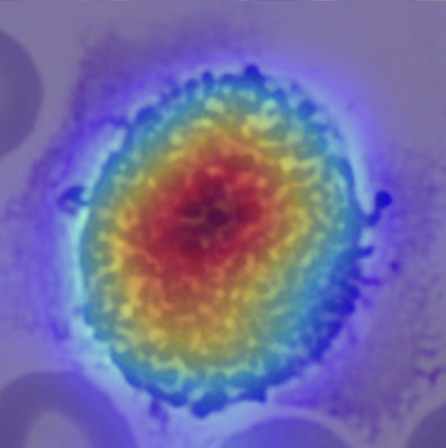} &
        \includegraphics[width=0.17\textwidth]{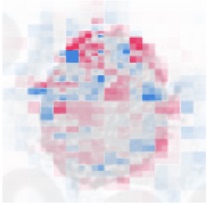} &
        \includegraphics[width=0.17\textwidth]{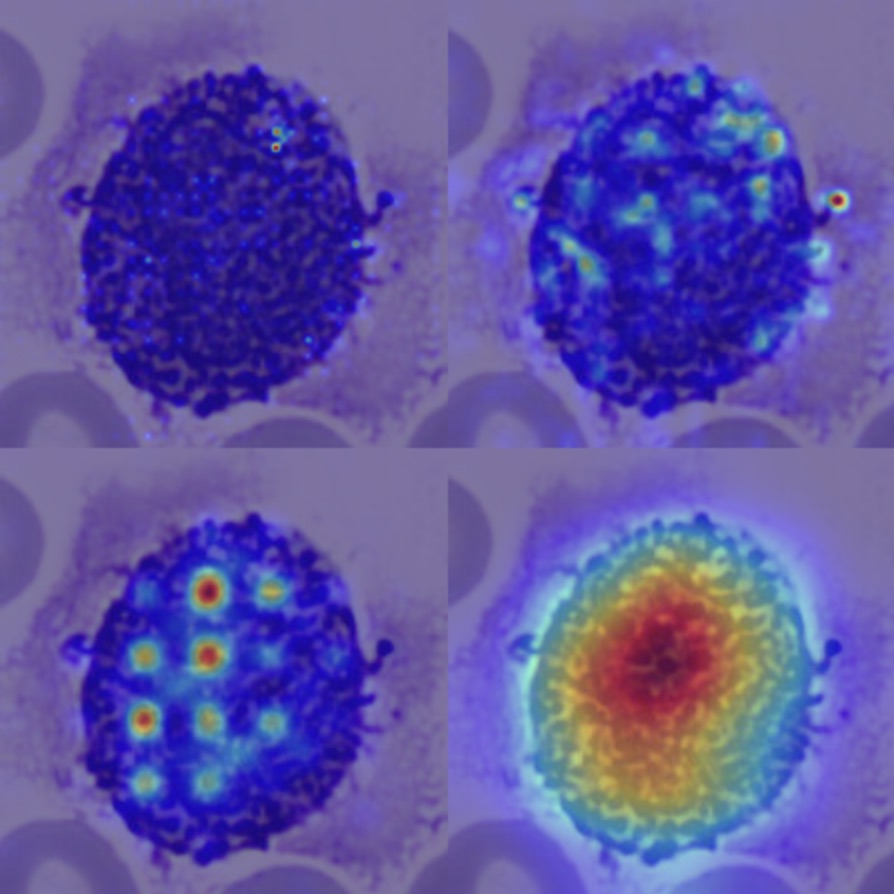} \\ \hline
        \rotatebox[origin=c]{90}{\makecell{\textbf{Monocyte}}} &
        \includegraphics[width=0.17\textwidth]{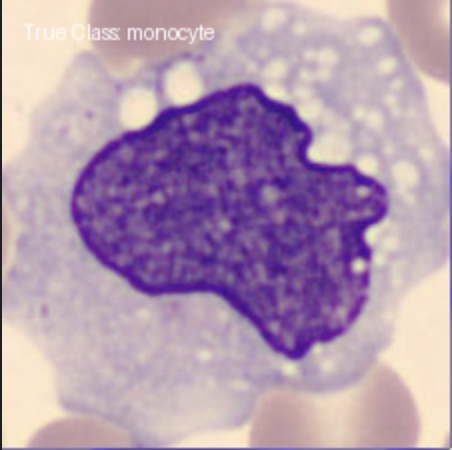} &
        \includegraphics[width=0.17\textwidth]{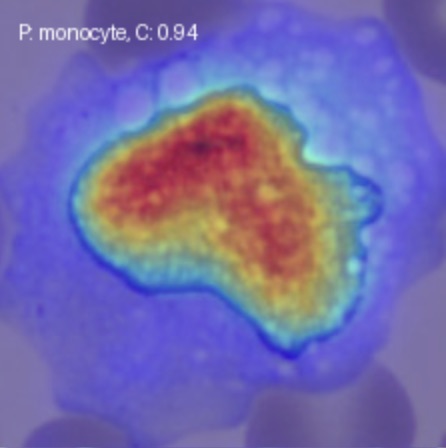} &
        \includegraphics[width=0.17\textwidth]{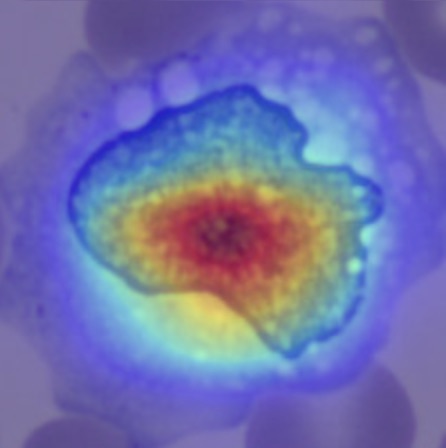} &
        \includegraphics[width=0.17\textwidth]{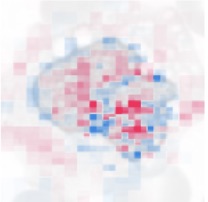} &
        \includegraphics[width=0.17\textwidth]{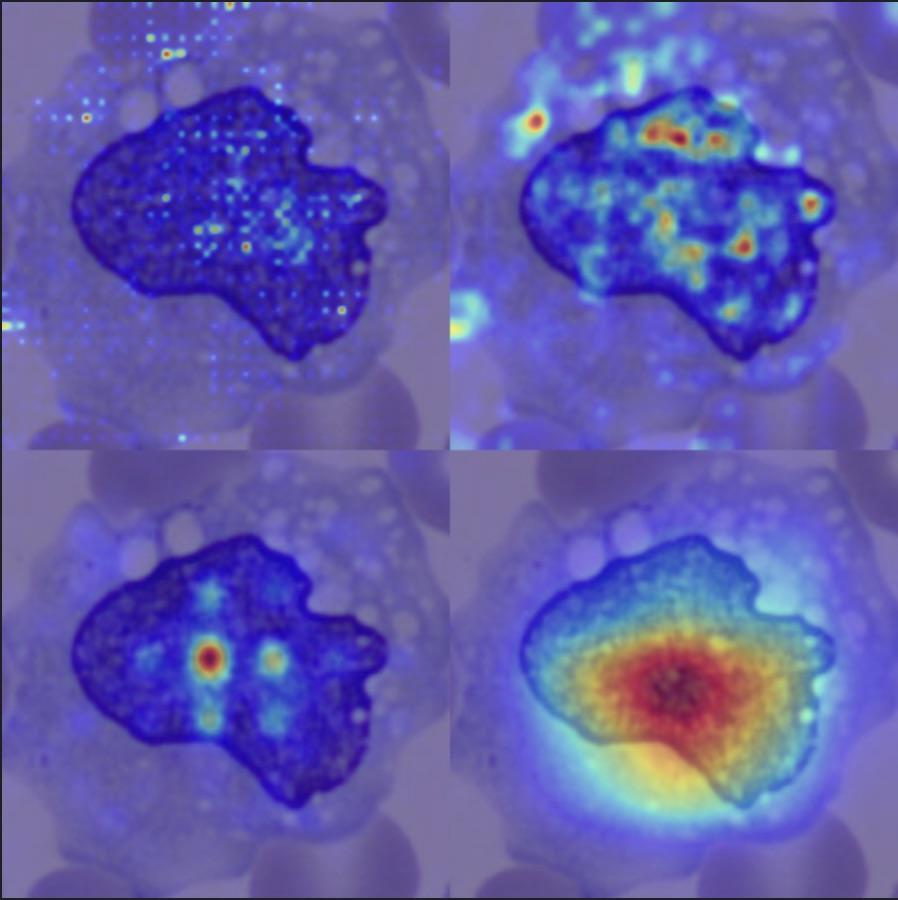} \\ \hline
        \rotatebox[origin=c]{90}{\makecell{\textbf{Neutrophil}}} &
        \includegraphics[width=0.17\textwidth]{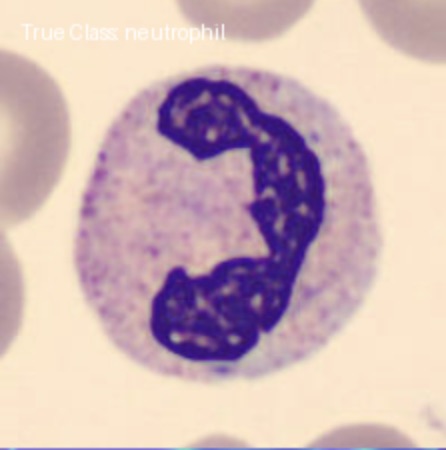} &
        \includegraphics[width=0.17\textwidth]{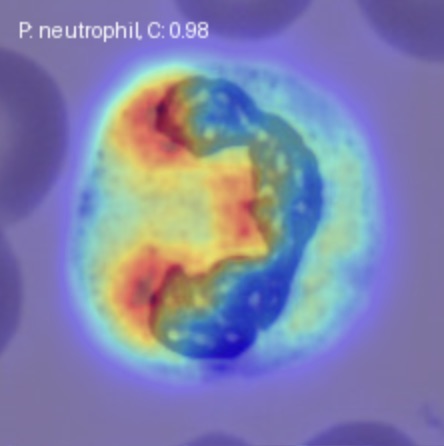} &
        \includegraphics[width=0.17\textwidth]{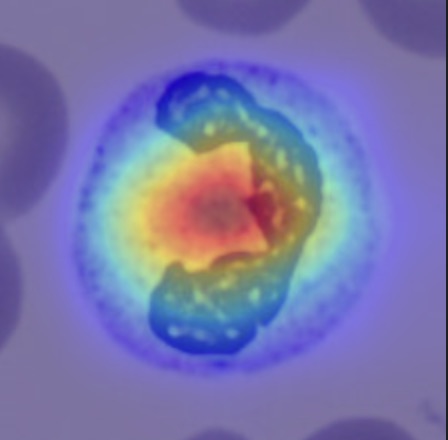} &
        \includegraphics[width=0.17\textwidth]{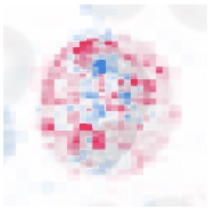} &
        \includegraphics[width=0.17\textwidth]{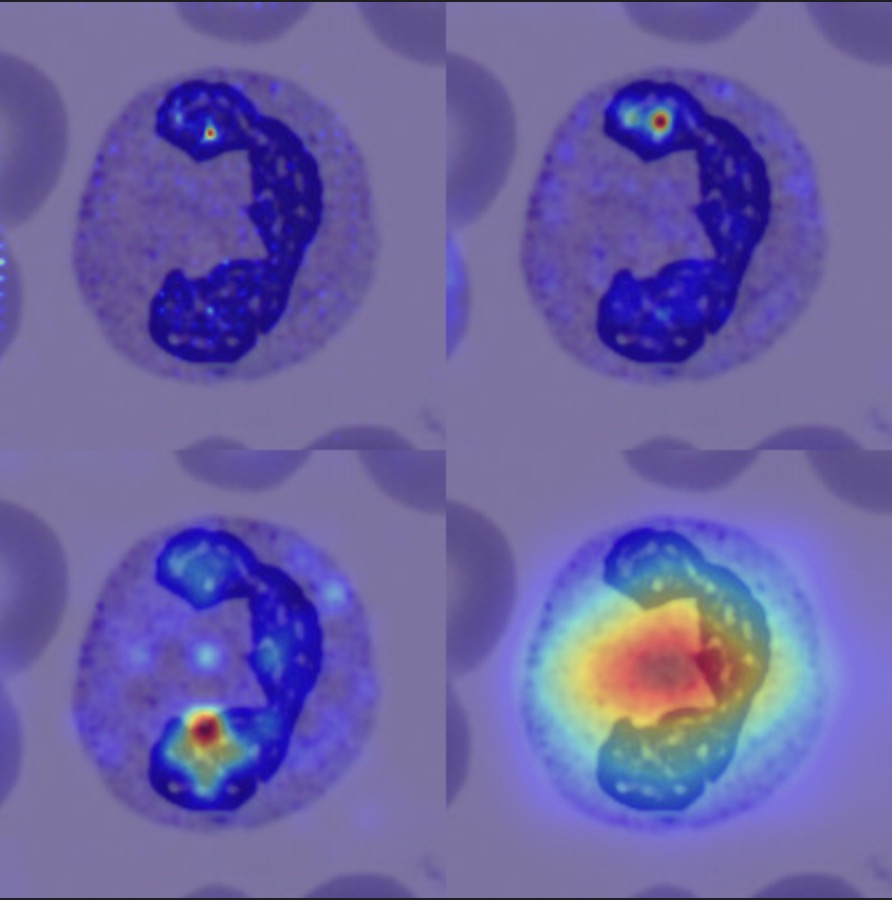} \\ \hline
        \rotatebox[origin=c]{90}{\makecell{\textbf{Immature}\\[-0.5em]\textbf{\hspace{2pt}Granulocytes}}} &
        \includegraphics[width=0.17\textwidth]{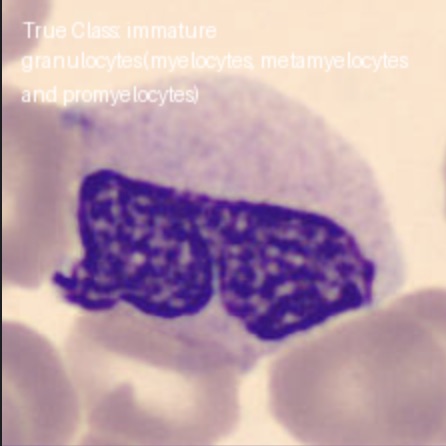} &
        \includegraphics[width=0.17\textwidth]{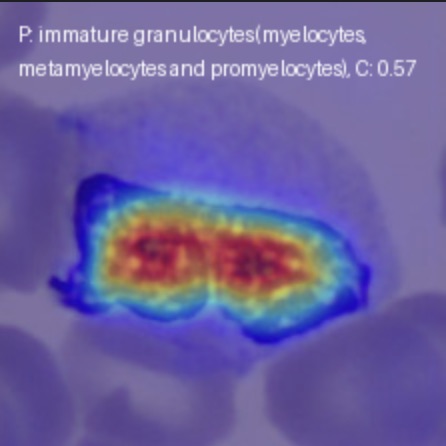} &
        \includegraphics[width=0.17\textwidth]{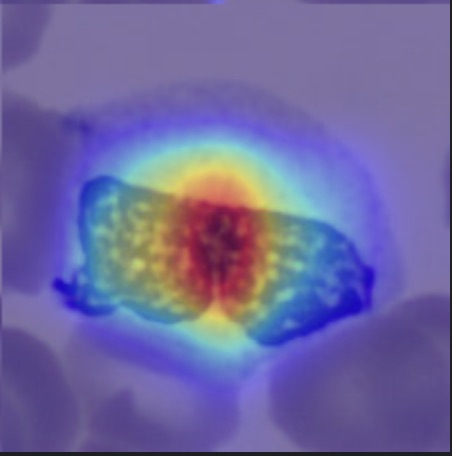} &
        \includegraphics[width=0.17\textwidth]{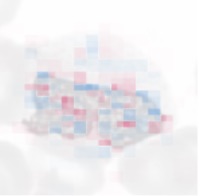} &
        \includegraphics[width=0.17\textwidth]{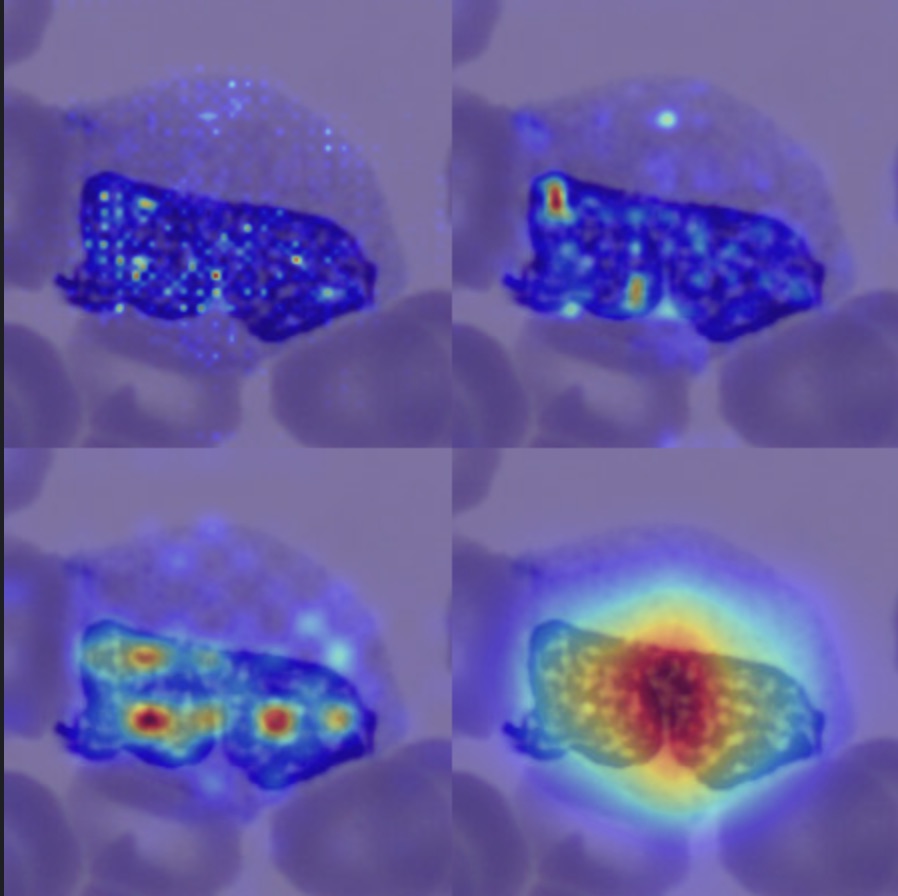} \\ \hline
    \end{tabular}
    \caption{Comparison of saliency maps for BloodMNIST. MHEX demonstrates its capability to capture fine-grained features.}
    \label{tab:bloodmnist}
\end{figure*}

\section{Training and Fine-Tuning Details}
\label{appendix:Training and Fine-Tuning Details}

\subsection{Pretraining on ImageNet1k}

For pretraining, we employ the ResNet-18 architecture as the backbone, integrating the \textbf{MHEX} modules into each residual connection. The detailed setup is as follows:
\begin{itemize}
    \item \textbf{Dataset:} ImageNet1k.
    \item \textbf{Architecture:} ResNet-18 with \textbf{MHEX} modules in residual connections.
    \item \textbf{Epochs:} 90 for pretraining from scratch, or 25 for fine-tuning from pretrained ResNet-18 weights. Both methods lead to similar performance results.
    \item \textbf{Batch Size:} 468.
    \item \textbf{Learning Rate:} Initialized at \(0.001\), decayed using a cosine annealing scheduler.
    \item \textbf{Optimizer:} AdamW with \(\beta = (0.9, 0.999)\) and a weight decay of \(0.01\).
    \item \textbf{Data Augmentation:} Random resized cropping, horizontal flipping, and TrivialAugmentWide.
    \item \textbf{Mixed Precision Training:} Automatic mixed precision (AMP) is used throughout both pretraining and fine-tuning to accelerate training and reduce memory consumption.
\end{itemize}

\subsection{Fine-Tuning on MedMNIST}

Fine-tuning is conducted on four datasets from MedMNIST (\textit{PathMNIST, BloodMNIST, TissueMNIST, OrganAMNIST}) with the following configuration:

\begin{itemize}
    \item \textbf{Datasets:}
    \begin{itemize}
        \item \textit{PathMNIST:} Multi-class classification of colon pathology images (9 classes, 107,180 samples).
        \item \textit{BloodMNIST:} Multi-class classification of blood cell images (8 classes, 17,092 samples).
        \item \textit{TissueMNIST:} Multi-class classification of kidney cortex images (8 classes, 236,386 samples).
        \item \textit{OrganAMNIST:} Multi-class classification of abdominal CT images (11 classes, 58,830 samples).
    \end{itemize}
    \item \textbf{Learning Rate:} \(1 \times 10^{-5}\).
    \item \textbf{Optimizer:} AdamW with a weight decay of \(0.01\).
    \item \textbf{Data Augmentation:} None.
    \item \textbf{Epochs:} 20.
\end{itemize}

\subsection{Fine-Tunin on BERT}

We utilize the pretrained BERT-base-uncased model available from Hugging Face\footnote{\url{https://huggingface.co/google-bert/bert-base-uncased}}. 

\begin{itemize}
    \item \textbf{Optimizer}: AdamW with \(\beta = (0.9, 0.999)\) and a weight decay of \(0.01\).
    \item \textbf{Learning Rate}: 2e-5
    \item \textbf{Batch Size}: 32.
    \item \textbf{Epochs}: 5.
    \item \textbf{Loss Function}: Cross-Entropy Loss.
\end{itemize}

Early stopping based on validation loss is employed to prevent overfitting. Training is conducted using an NVIDIA RTX 4090 GPU to expedite convergence.

\section{Guidelines}
\label{appendix:Guidelines}
To facilitate the effective application of our proposed \textbf{Multi-Head Explainer (MHEX)} framework, we offer the following guidelines based on our experimental observations:

\begin{itemize}[left=0pt]
    \item \textbf{Training Methods}: In Section~\ref{sec:Deep Supervision}, we present two training approaches. If the primary focus is on enhancing accuracy, we recommend utilizing the pretraining method. This approach can improve accuracy by approximately 0.5\% to 1\% on the ImageNet1k dataset. Although early layers contribute less to predicting the correct class, our observations indicate that they help in eliminating certain incorrect classes, thereby refining the model's predictive performance.
    
    \item \textbf{Residual Connection Placement}: While theoretically, the MHEX framework can be integrated into any residual connection, we advise positioning it at the information bottleneck points. Specifically, our experiments demonstrate that deploying MHEX on the downsampling residual connections of ResNet architectures yields superior explanations and accuracy, particularly in the early layers. 
    
    \item \textbf{Necessity of Attention Gates}: The inclusion of \textbf{Attention Gates} is not strictly mandatory; however, excluding them results in a significant reduction in the confidence of lower-level predictions. We experimented with separating the Attention Gate by setting \( W_{\text{equiv}} = W_2 \). This modification occasionally produced cleaner saliency maps but at the cost of substantially decreased prediction confidence. Therefore, retaining the Attention Gate is recommended to maintain a balance between saliency map quality and prediction reliability.
    
    \item \textbf{Deployment Flexibility}: There is no requirement to remove the original prediction heads or [CLS] tokens when integrating MHEX. Additionally, it is not necessary to deploy MHEX across every residual connection. The decision to include MHEX modules should be tailored to the specific requirements of the task at hand. 
    
    \item \textbf{Application to Other Networks}: We believe that MHEX is applicable to a variety of neural network architectures beyond CNNs and Transformers. For instance, in Graph Neural Networks (GNNs), MHEX should function effectively by treating graph nodes analogously to tokens in transformer-based models. However, it is important to note that standard GNN architectures, such as Graph Convolutional Networks (GCNs) \cite{kipf2016semi} and Graph Attention Networks (GATs) \cite{velivckovic2017graph}, typically lack residual connections. Therefore, a key consideration is how to introduce residual links within these frameworks to seamlessly integrate MHEX. Additionally, MHEX can also be applied to medical image segmentation tasks, particularly when used with powerful architectures like UNet and UNet-Transformer \cite{ronneberger2015u,petit2021unettransformerselfcross}. Given the different nature of segmentation tasks, the primary modification involves replacing the equivalent matrix (\( W_{\text{equiv}} \)) with an equivalent convolution (\( \text{Conv}_{\text{equiv}} \)), allowing MHEX to better adapt to the spatial features inherent in these tasks.
\end{itemize}




\end{document}